\documentclass[journal]{IEEEtran}

\usepackage{graphicx}
\usepackage{comment}
\usepackage{amsmath,amssymb} 
\usepackage{color}
\usepackage{amsmath}
\usepackage{footnote}
\usepackage{xcolor,soul,framed} 
\usepackage{xspace}
\usepackage{array}
\usepackage{eqparbox}
\usepackage{url}
\usepackage{longtable}
\usepackage{lipsum}
\usepackage{blindtext}
\usepackage{makecell}
\usepackage{mathtools}
\usepackage{commath}
\usepackage{multirow}
\usepackage{tabularx}
\usepackage[normalem]{ulem}
\usepackage{xspace}
\usepackage{algorithm}
\usepackage{algpseudocode}
\usepackage{amsfonts}
\usepackage{subcaption}
\usepackage{xcolor}
\usepackage[symbol]{footmisc}
\newcommand{\cmark}{\ding{51}}%
\newcommand{\xmark}{\ding{55}}%

\usepackage{amssymb}
\usepackage{pifont}

\usepackage{enumitem}

 \setlist[itemize]{leftmargin=*}

\begin{document}
%
\title{CapsField: Light Field-based Face and Expression Recognition in the Wild using Capsule Routing}

\author{Alireza Sepas-Moghaddam \IEEEmembership{Member, IEEE}, Ali Etemad \IEEEmembership{Senior Member, IEEE}, Fernando Pereira \IEEEmembership{Fellow, IEEE}, and Paulo Lobato Correia, \IEEEmembership{Senior Member, IEEE}
}


\markboth{ }%
{Shell \MakeLowercase{\textit{et al.}}: Bare Demo of IEEEtran.cls for IEEE Journals}


\maketitle
\begin{abstract}

Light field (LF) cameras provide rich spatio-angular visual representations by sensing the visual scene from multiple perspectives and have recently emerged as a promising technology to boost the performance of human-machine systems such as biometrics and affective computing. Despite the significant success of LF representation for constrained facial image analysis, this technology has never been used for face and expression recognition \textit{in the wild}. In this context, this paper proposes a new deep face and expression recognition solution, called CapsField, based on a convolutional neural network and an additional capsule network that utilizes dynamic routing to learn hierarchical relations between capsules. CapsField extracts the spatial features from facial images and learns the angular part-whole relations for a selected set of 2D sub-aperture images rendered from each LF image. To analyze the performance of the proposed solution in the wild, the first in the wild LF face dataset, along with a new complementary \textit{constrained} face dataset captured from the same subjects recorded earlier have been captured and are made available. A subset of the in the wild dataset contains facial images with different expressions, annotated for usage in the context of face expression recognition tests. An extensive performance assessment study using the new datasets has been conducted for the proposed and relevant prior solutions, showing that the CapsField proposed solution achieves superior performance for both face and expression recognition tasks when compared to the state-of-the-art.

\begin{IEEEkeywords}
Face Recognition, Expression Recognition, Light Field, Face Dataset, Deep Learning, Capsule Routing.
\end{IEEEkeywords}
\end{abstract}

\section{Introduction} \label{sec:intro}
Face is the most common characteristic used by humans to perform personal identity \cite{jain,tax} and expression recognition \cite{6940284}. Following the emergence of the first automatic facial image analysis system in the late 1960’s \cite{first}, this field has attracted considerable research efforts leading to incredible progresses. Despite the recent advances in face and expression recognition, mostly thanks to the advancements in deep neural networks \cite{deepsurvey, kittler, deepemotion}, highly accurate recognition results are still not achievable for some specific conditions. This is mostly the case when poor quality data is available, for instance due to the capture of face images in uncontrolled settings, commonly referred to as face recognition in the wild, where facial images are affected by multiple variations, notably in resolution, background, expression, pose, illumination, and occlusions, among others \cite{jain2}.

The emergence of new imaging sensors has opened new frontiers and also brought new challenges for face image analysis systems \cite{sensor}. Lenslet light field cameras, hereafter referred only as \textit{Light Field (LF)} cameras, have recently come into prominence as they are able to simultaneously capture the intensity of light rays coming from multiple directions in space \cite{lensletLF,LF}. An LF camera ``sees'' the visual scene from multiple angles or perspectives, allowing to render a set of 2D \textit{Sub-Aperture (SA)} images, each corresponding to a 2D image for a different viewpoint of the observed scene. The set of rendered 2D SA images forms a multi-view SA array which offers intra-view/spatial (within each view) and inter-view/angular (across views) correlations that can be effectively exploited for various visual analysis tasks, notably facial image analysis \cite{LFsurvey}. {In this context, the main challenge of LF-based face and emotion recognition is how to exploit both the intra-view and inter-view relationships to improve the performance of the recognition solutions. This is the recognition paradigm considered in this paper. The LF-based face and emotion recognition can also be formulated as an image set based recognition problem \cite{imageset}, where matching is performed between image sets instead of single images.}

Despite the significant success of the LF representation for constrained face recognition \cite{ryrb13,rryb13, rrb16, icip, ear, MLSP, CSVT, joint} and expression recognition \cite{ACII,ICASSP}, 
this type of visual representation has never been used for face and expression recognition in the wild, where the added value of the LF information may be more critical since the conditions are more extreme and unconstrained.
In this context, this paper proposes to exploit the richer spatio-angular representation acquired with a LF camera, to improve the performance of face and expression recognition systems in the wild. In this paper, \textit{‘in the wild’} refers to facial images captured under several {unconstrained acquisition conditions such as random lighting conditions, shading, backgrounds, environments, poses, distances, etc.,} with minimal or no user cooperation. The key contributions and technical novelty of this paper in the context of LF face and expression recognition in the wild can be described as follows:

\begin{itemize}

    \item \textbf{CapsField Solution:} A new deep learning solution, called CapsField, is proposed for both face and expression recognition in the wild. CapsField combines convolutional and capsule networks \cite{capsule} for exploiting not only spatial information but also the angular information available in LF images. A few solutions based on capsule networks have recently been proposed in the context of facial image analysis, including age and gender \cite{CapFace1}, and expression \cite{CapFace2} recognition, showing that capsule networks are well suited to deal with face pose variations. CapsField goes one step further, for the first time exploiting dynamic routing between capsules for learning the angular variations across the different viewpoints captured in LF images. Additionally, CapsField assigns higher weights to the more relevant features while ignoring the spurious dimensions, to increase its robustness against face appearance and environmental changes.

    \item \textbf{In the wild face data collection:} The captured dataset, named Light Field Faces in the Wild (LFFW), addresses a major weakness in the face and expression recognition research domains, which is the lack of publicly available LF facial images captured in the wild. To the best of the authors’ knowledge, no similar dataset is available, even for private usage. The proposed LFFW dataset consists of LF images captured under several unconstrained acquisition variations, in both indoor and outdoor environments, at different locations, and from different distances. {During the acquisition, there were no pre-defined protocol or any stimulus used to induce the desired expression}. Additionally, a subset of the LFFW dataset includes facial images with different expressions, which has also been annotated to be used in the context of LF-based facial expression recognition.

    \item \textbf{Constrained face data collection:} This paper also makes available a second LF dataset, called Light Field Face Constrained (LFFC), acquired in constrained conditions, with the same subjects present in LFFW. 
    The complementary LFFC dataset was acquired between 1 day and 3 years prior to the LFFW images, allowing to study the constrained versus in the wild cross-dataset generalization ability of face recognition solutions. The two complementary face datasets (LFFW and LFFC) offer a unique set of facial data, facilitating a wide range of research possibilities related to face analysis. Both datasets, along with the annotation metadata, will be publicly available to the research community.
  
    \item \textbf{Benchmarking:} Finally, this paper provides an extensive performance assessment of the proposed face and expression recognition solutions regarding the most competitive state-of-the-art solutions. As new datasets along with novel evaluation protocols are proposed in this paper, the authors had to re-implemented all the benchmarking solutions. The obtained results show that the proposed CapsField solution outperforms the state-of-the-art solutions by a significant margin for both face and expression recognition tasks, for the new face datasets.
    
\end{itemize}

The rest of this paper is organized as follows: Section II briefly reviews the basic concepts and the added value of LF imaging for face recognition and provides a comprehensive review of recent advances in LF-based face and expression recognition solutions and datasets to better understand the technological landscape in this field.
The proposed CapsField solution is presented in Section III and the new face datasets are described in Section IV. Section V presents an extensive performance evaluation for the proposed and state-of-the-art solutions using a novel evaluation framework addressing varied and challenging recognition tasks. Finally, Section VI concludes the paper.

\section{Background and Related Work}
This section starts with an overview of light field imaging basics, followed by a LF-based face and expression recognition state-of-the-art review, and finally overviews the main face datasets available, including the LF ones.

\subsection{Light Field Imaging Basics}\label{sec:related_work}
LF imaging has been researched for more than a century \cite{integral}. In 1908, Lippmann used small and closely spaced circular lenses to record different perspectives of a scene, so that an image could be observed from a selected perspective through an array of lenses \cite{lightfield}. In 1936, Gershun coined the term \textit{light field}, referring to the amount of light traveling in all directions through every point in space \cite{ab91}. In 1991, Adelson proposed the Plenoptic function \textit{P(x,y,z,t,$\lambda$,$\theta$,$\phi$)}, describing the information carried by light rays at any point in 3D space (\textit{x,y,z}), for all directions ($\theta$,$\phi$) and wavelengths ($\lambda$), along time (\textit{t}). 
The static 4D LF \cite{LF}, \textit{L(x,y,u,v)}, also known as Lumigraph \cite{lumi}, proposed a representation based on the intersection points of the light rays with two parallel planes \cite{d14}. 

\begin{figure}[!t]
\centering
\includegraphics[width=0.85\columnwidth]{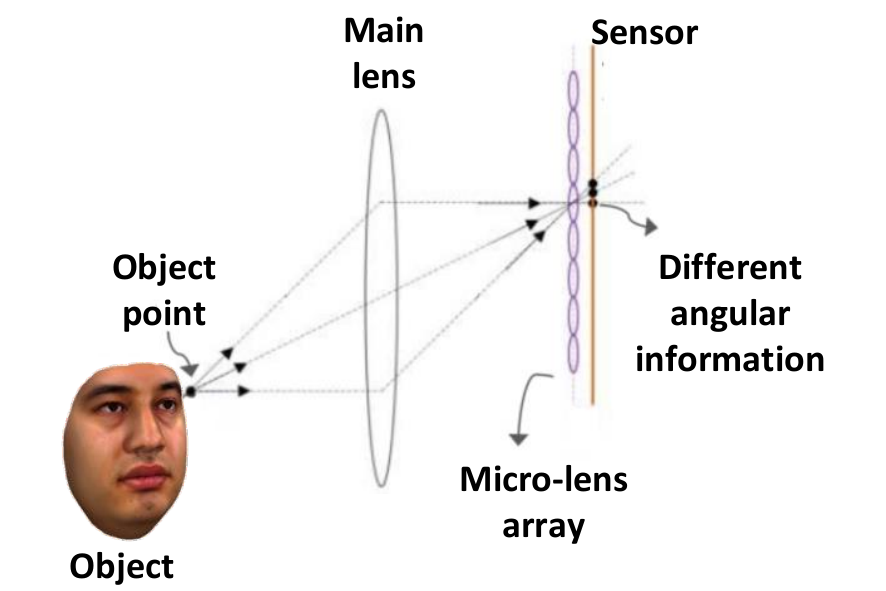}
\caption{{Lenslet light field imaging system.}}
\label{fig:LFCapt}
\end{figure}

\begin{figure}[!t]
\centering
\includegraphics[width=1\columnwidth]{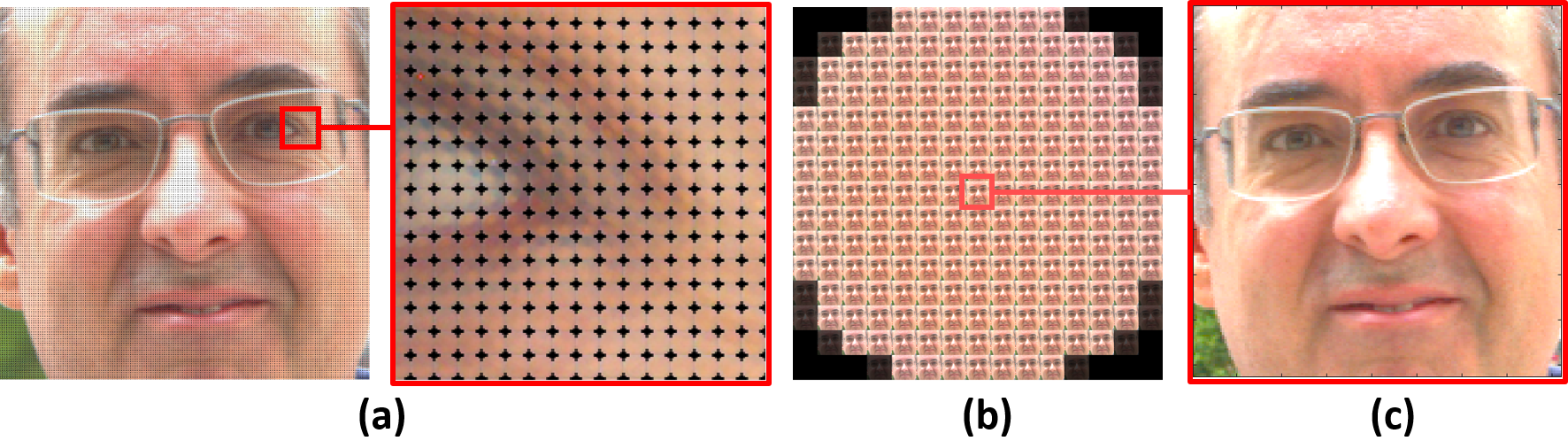}
\caption{Light field representation: a) sample micro-images, after colour demosaicing, also with zoom; b) corresponding multi-view SA array; c) corresponding 2D central SA image.}
\label{fig:LF}
\end{figure}

For capturing LF images, the type of visual data adopted in this work, two main practical setups are currently available, notably using a high density array of cameras or a lenslet LF camera. 
In a lenslet LF camera, {as illustrated in Figure \ref{fig:LFCapt},} a micro-lens array, consisting of a set of micro-lenses, is placed at the focal plane of the main lens at a given distance from the photo sensor \cite{lensletLF}. This allows 
for the incoming light to be split based on its direction, and projected onto the photo sensor area. Each micro-lens can be thought of as a small camera, capturing light rays from a specific point in the scene, thus acquiring a so-called micro-image \cite{ps16}. A sample of a full set of LF color demosaiced micro-images is shown in Figure \ref{fig:LF}(a). A set of SA images, corresponding to different viewpoints of the scene can then be rendered from the micro-images. The set of rendered SA images can be considered as a multi-view SA array. Examples of a multi-view SA array and the corresponding 2D central SA image are shown in Figure \ref{fig:LF}(b) and Figure \ref{fig:LF}(c), respectively. Since each 2D SA image `sees' the scene from a slightly different viewpoint, the full set of 2D images provides angular variations of the scene, which is a distinctive characteristic of LF imaging. The LFFW dataset proposed in this paper was captured with a Lytro Illum camera \cite{Lytro}, which multi-view SA array includes $15\times15$ SA images, each SA image with a spatial resolution of $625\times434$ pixels.

In this context, there are two different types of correlation available within a multi-view SA array: i) the spatial intra-view correlation within views; and ii) the angular inter-view correlation between different views. Therefore, LF images open the possibility for spatio-angular information to be exploited, allowing to boost the face and expression recognition performance by: i) \textit{a posteriori} refocusing to improve the quality of the face region(s) that may be out-of-focus during acquisition; ii) exploiting the disparity for representing variations associated with different light directions captured in LF images; and iii) exploiting depth for providing geometric information about the position and shape of facial components.


\subsection{Light Field-based Face Recognition Solutions}
Recently, several face recognition solutions using LF images have been proposed. This section reviews this category of face recognition solutions, which are able to exploit the availability of LF sensors and data. The main characteristics of the most notable recognition solutions in this area are summarized in Table \ref{tab:Method}, including the adopted feature extraction approach, classifier, exploited LF capability (as introduced in Section II), the type of LF data, and the datasets used by each solution for reporting performance results. For comparison, Table \ref{tab:Method} also includes the characteristics of the CapsField solution proposed in this paper. The solutions summarized in the table are reviewed here, notably grouped based on the feature extraction approaches considered.

\begin{table*}[]

\centering
\setlength\tabcolsep{5 pt}
\footnotesize
\centering
\caption{Overview of main available light field-based face recognition solutions.}

\begin{tabular}{l|l|l|l|l|l|l|l}

\hline
\textbf{Solution}            & \textbf{Year} & {\vtop{\hbox{\strut \textbf{Feature Extract.}}\hbox{\strut \textbf{Approach}}}}        & {\vtop{\hbox{\strut \textbf{Feature Extraction}}\hbox{\strut \textbf{Method}}}} & \textbf{Classifier} & \textbf{Light Field Capability}                    & \textbf{Data}                & \textbf{Dataset} \\\hline 
\hline
LF Face \cite{ryrb13}         & 2013 &  Visual Descriptor                           & LBP                       & NN        & Depth Computation                         & LF 2D Rendered & Private \\ 
Multi-Face LF \cite{rryb13}   & 2013 &  Visual Descriptor                           & LBP; LG ﬁlter             & SRC        & \textit{A Posteriori} Refocusing                   & LF 2D Rendered & LiFFID  \\
Super Res. LF \cite{rryb16}   & 2013 &  Visual Descriptor                           & LBP                       & SCR        & \textit{A Posteriori} Refocusing                   & LF 2D Rendered & LiFFID  \\
Face-Iris MF LF \cite{rrb16} & 2016 &  Visual Descriptor                           & HOG; LBP; BSIF  & SRC        & \textit{A Posteriori} Refocusing                   & 2D Rendered & LiFFID  \\
DM LF  \cite{sfc16}          & 2016 & Visual Descriptor                           & LFHOG                     & SVM        & Depth Computation                         & M-V SA Array        & Private \\
LFLBP \cite{icip}          & 2017 &  Visual Descriptor                           & LFLBP                     & SVM        & Disparity Exploitation                    & M-V SA Array        & LFFD    \\
LFHG \cite{ear}            & 2018 &  Visual Descriptor                           & HOG; LFHDG                & SVM        & Disparity Exploitation                    & M-V SA Array        & LFFD    \\ 
VGG-D3 \cite{MLSP}          & 2018 & Deep Nets                  & VGG                       & SVM        &  Disparity \& Depth Exploit.   & M-V SA Array        & LFFD    \\
VGG+ Conv-LSTM \cite{CSVT}  & 2019 & Deep Nets                  & VGG;  Conv-LSTM           & Softmax    & Disparity Exploitation                    & M-V SA Array        & LFFD    \\ 
VGG+ GLF-LSTM \cite{joint}   & 2019 & Deep Nets                  & VGG; GLF-LSTM             & Softmax    & Disparity Exploitation                    & M-V SA Array        & LFFD    \\ 
VGG+ SLF-LSTM \cite{joint}   & 2019 & Deep Nets                  & VGG; SLF-LSTM             & Softmax    & Disparity Exploitation                    & M-V SA Array        & LFFD    \\
VGG+ SeqL-LSTM \cite{joint}  & 2019 & Deep Nets                  & VGG; SeqL-LSTM            & Softmax    & Disparity Exploitation                    & M-V SA Array        & LFFD\\ \hline 
Prop. ResNet+Capsule  & 2020 & Deep Nets                  & {\vtop{\hbox{\strut ResNet50 +}\hbox{\strut Capsule Network}}}   & Softmax    & Disparity Exploitation                    & M-V SA Array        & {\vtop{\hbox{\strut LFFW;}\hbox{\strut LFFC}}}\\ 
\hline
\end{tabular}
\label{tab:Method}
\end{table*}


\subsubsection{Visual Descriptor-based Solutions}
The first group of visual descriptor-based solutions relies on the \textit{a posteriori} refocusing capability to improve the quality of out-of-focus facial regions. In \cite{ryrb13} a wavelet energy method is proposed for selecting the best focus plane. Then, the Local Binary Pattern (LBP) descriptor is applied to extract features and a Nearest Neighbor (NN) classifier is used for performing recognition. Another solution uses a resolution enhancement scheme \cite{rryb13} based on a discrete wavelet transform, to capture the components with highest frequencies from different refocused images. This solution creates after an all-in-focus face image to be input to an LBP descriptor. In \cite{rryb16}, the recognition of multiple faces placed at different distances with an all-in-focus image rendered from the LF is investigated. An LBP descriptor is then applied for feature extraction and a Sparse Reconstruction Classifier (SRC) is used for classification. In \cite{rrb16}, the LF-based face recognition solution renders a set of refocused images using two different approaches. The first selects the best refocused image for recognition while the second combines the refocused images for creating a super-resolved image. Different local descriptors including Histogram of Oriented Gradient (HOG), LBP, Center-Symmetric LBP (CSLBP), and Binarized Statistical Image Features (BSIF) are used to extract features for classification.    

The second group of visual descriptor-based solutions exploits the depth information that can be estimated from an LF image. These solutions explore the geometry of the facial components, notably their positions and shapes. In \cite{sfc16}, a depth map is exploited to extract discriminative HOG features used as input to a linear Support Vector Machine (SVM) for classification.

The last group of visual descriptor-based solutions relies on exploiting the LF disparity information. In \cite{icip}, a face recognition solution is proposed based on a visual descriptor named Light Field LBP (LFLBP), exploiting the richer spatio-angular LF information. LFLBP has two main components, the spatial LBP and the angular LBP, capturing not only the usual spatial information but also the angular information available in a set of SA images. Another solution is proposed in \cite{ear} based on a the Light Field Histogram of Gradients (LFHG) descriptor, which fuses the conventional HOG with a Light Field Histogram of Disparity Gradients (LFHDG) descriptor, to consider both the orientation and magnitude information.

\subsubsection{Deep Learning-based Solutions}
Recognizing the importance of deep learning in biometric recognition, an LF face recognition solution has been designed based on a VGG 2D+Disparity+Depth (VGG-D3) fused deep descriptor \cite{MLSP}. The VGG-D3 description is formed by concatenating descriptions extracted from 2D images as well as disparity and depth maps using the VGG-16 descriptor trained on the VGG-Face 1 dataset \cite{pvz15}. This was the first solution to adopt a fused deep CNN network to exploit the complementary LF information, including 2D texture and the corresponding disparity and depth maps. Another face recognition solution is based on a double-deep descriptor, so-called VGG + Conventional Long Short Term Memory (Conv-LSTM) \cite{CSVT}. This solution exploits the multi-perspective LF information While the VGG-D3 \cite{MLSP} processes only the information available in the 2D central SA image. The double-deep descriptor extracts spatio-angular LF dependencies using VGG and LSTM networks in sequence, to provide a more powerful description for face recognition. The solution in \cite{joint} proposes three novel LSTM cell architectures to jointly learn from LF horizontal and vertical parallaxes. The three cell architectures perform: i) Gate-Level Fusion LSTM (GLF-LSTM); ii) State-Level Fusion LSTM (SLF-LSTM); and iii) Sequential Learning LSTM (SeqL-LSTM). These architectures create richer spatio-angular LF descriptions for face recognition and have been integrated into an end-to-end framework, where a VGG network feeds the LSTM networks composed of the novel LSTM cell architectures.

In summary, given the nature of the available face recognition solutions and the way they were assessed, notably using different datasets and evaluation protocols, it is difficult to compare the various solutions and even more difficult to estimate how these solutions would perform in unconstrained operational conditions. Since more recent research has shifted towards in the wild face recognition, this paper addresses this shortcoming by presenting an extensive benchmarking study, performed on the LFFW dataset proposed in this paper.

\subsection{Light Field-based Expression Recognition Solutions}
There are currently two solutions available for exploiting additional LF information for face expression recognition. Inspired by \cite{CSVT}, a deep learning spatio-angular fusion framework was adopted in \cite{ACII}, to model both the spatial and angular information with VGG-16 and LSTM recurrent networks. In \cite{ICASSP}, the deep framework proposed in \cite{ACII} was extended by exploring both forward and backward angular relations using bidirectional LSTM. Additionally, an attention mechanism selectively focuses on the most important spatio-angular features, thus learning the final features more effectively. 

\subsection{{Sampling Schemes and Fusion Strategies}}

{In order to exploit the angular dependencies between LF views, the views should be arranged in the form of a pseudo-video sequence to feed an LSTM network. In \cite{CSVT}, 11 different sampling schemes were considered to select and scan the set of SA images, notably varying their number, position, and scanning order. The results showed that capturing angular information along the horizontal and vertical directions leads to better average performance when compared to other sampling schemes, notably due to the consideration of larger disparities. The results also showed that the score-level fusion of horizontal and vertical angular classification models leads to additional improvements.}

{An alternative approach to the score-level fusion strategy is to jointly learn the horizontal and vertical view sequences, as done by the GLF-LSTM and SLF-LSTM cell architectures proposed in \cite{joint} (introduced in Section II.B). Additionally, there are other alternative solutions, not originally designed for the same task, but which are flexible enough to be adopted for simultaneously learning from the horizontal and vertical view sequences. 
Examples are the Spatio-Temporal LSTM (ST-LSTM) \cite{LSTMV1} and the Dual-Sequence LSTM (DS-LSTM) cell architecture \cite{LSTMV2}, which were originally designed to fuse different sequences for activity and speech recognition, respectively. These alternative LSTM cell architectures will be considered for benchmarking purposes, when comparing to the proposed solution in this paper.}

\subsection{Face Datasets: \textit{Status Quo}}
There are currently over 150 publicly available face datasets \cite{facedb}. The characteristics of the more notable face datasets are presented in Table \ref{tab:DB}, highlighting variations in terms of acquisition conditions, expressions, poses, and occlusions. For easier reading, the datasets are sorted according to their release date. For comparison, Table \ref{tab:DB} also includes the characteristics of the LFFW and LFFC datasets proposed in this paper.

\begin{table*}[!t]
\centering
\setlength\tabcolsep{1.5pt}
\footnotesize
\centering
\caption{Overview of main face datasets with relevant characteristics.}
\setlength\tabcolsep{5pt}
\begin{tabular}{ l | l| l| l| l| l| l| l| l| l| l| l| l}
\hline
 \multicolumn{1}{ c |}{\textbf{Dataset Name}}& \multicolumn{1}{ c |}{\textbf{Year}}& \multicolumn{1}{ c| }{\textbf{\# of 2D Images}}& \multicolumn{1}{ c| }{\textbf{In the Wild?}}&\multicolumn{1}{ c| }{\textbf{Image Type}}&\multicolumn{1}{ c| }{\textbf{Sensor}}& \multicolumn{7}{ c }{\textbf{Face Variation}}  \\ 
\hline
    & & & & & & Time & Env. & Dist. & Light & Poses & Exp. & Occl.\\ \hline\hline
    AR \cite{ar98} & 1998& $\approx$ 4k & No & Color & 2D & \cmark & \xmark & \xmark & \cmark & \xmark & \cmark & \cmark\\
    FERET \cite{pmrr00} & 2003 & $\approx$ 14k & No & Gray/Color & 2D &\cmark & \xmark & \xmark & \cmark & \cmark & \cmark & \cmark\\
    LFW \cite{hrbl07} & 2007 & $\approx$ 13k & Yes & Color & 2D &\cmark & \cmark & \cmark & \cmark & \cmark & \cmark & \cmark\\
    Multi-PIE \cite{gmckb10} & 2009 & $\approx$ 750k & No & Color & 2D &\xmark & \xmark & \xmark & \cmark & \cmark & \cmark & \xmark\\
    MOBIO \cite{m12} & 2010 & $\approx$ 30k & Yes & Color & 2D &\cmark & \cmark & \cmark & \cmark & \cmark & \cmark & \cmark\\
    YouTube Faces \cite{whm11} & 2011 &	$\approx$ 606k & Yes & Color & 2D Video &\cmark & \cmark & \cmark & \cmark & \cmark & \cmark & \cmark\\
    SCface \cite{gdg11} & 2011 & $\approx$ 4k & No & Color/Infra. & 2D; Infra.&\xmark & \xmark & \xmark & \cmark & \cmark & \cmark & \xmark\\
    BU-3DFE \cite{zyc14} & 2013 & $\approx$ 370k & No & Color & 2D+Depth &\xmark & \xmark & \xmark & \xmark & \cmark & \cmark & \xmark\\
    Kinect Face DB \cite{mkg14} & 2014 & $\approx$ 2k & No & Color & 2D+Depth &\cmark & \xmark & \xmark & \cmark & \cmark & \cmark & \cmark\\
    PIPA \cite{zptfb15} & 2015 & $\approx$ 63k & Yes & Color & 2D &\cmark & \cmark & \cmark & \cmark & \cmark & \cmark & \cmark\\
    VGG-Face 1 \cite{pvz15} & 2015 & $\approx$ 2.6 M & Yes & Color & 2D &\cmark & \cmark & \cmark & \cmark & \cmark & \cmark & \cmark\\
    LiFFID \cite{rrb16} & 2016 & $\approx$ 107k & No & Gray & LF &\xmark & \xmark & \xmark & \cmark & \cmark & \cmark & \xmark\\
    LFFD \cite{iwbf} & 2016 & $\approx$ 900k & No & Color & LF &\cmark & \xmark & \xmark & \cmark & \cmark & \cmark & \cmark\\
    MegaFace \cite{nk17} & 2017 & $\approx$ 4.7M & Yes & Color & 2D &\cmark & \cmark & \cmark & \cmark & \cmark & \cmark & \cmark\\
    DFW \cite{kssvrc18} & 2018 & $\approx$ 11k & Yes & Color & 2D &\cmark & \cmark & \cmark & \cmark & \cmark & \cmark & \cmark\\
    VGG-Face 2 \cite{csxpz18} & 2018 & $\approx$ 3.6M & Yes & Color & 2D &\cmark & \cmark & \cmark & \cmark & \cmark & \cmark & \cmark\\\hline
    Proposed LFFW & 2019 & $\approx$ 429k & Yes & Color & LF &\cmark & \cmark & \cmark & \cmark & \cmark & \cmark & \cmark\\
    Proposed LFFC & 2019 & $\approx$ 238k & No & Color &	LF &\xmark & \xmark & \xmark & \cmark & \cmark & \cmark & \cmark\\

    \hline
\end{tabular}
\label{tab:DB}
\end{table*}


Since LF imaging is a relatively new technology, to the best of the authors' knowledge, only two face datasets have been made publicly available.  {The Light Field Face and Iris Database (LiFFID) \cite{rrb16} was the first face dataset to include images acquired with a LF camera for facial recognition purposes. It includes a set of 2D greyscale images, focused at different depths, rendered from the LF content acquired using a first generation, lower resolution, Lytro lenslet camera. However, LiFFID does not include the raw LF images, which is a major limitation for many research applications.} The IST-EURECOM Light Field Face Database (LFFD) \cite{iwbf} was the second LF face dataset made available. This dataset includes raw and rendered data from 100 subjects, with 20 LF samples per person, where each LF sample includes 225 2D SA images, captured by a second generation Lytro ILLUM lenslet camera, in two separate acquisition sessions. The images are captured in a controlled acquisition setup with different facial variations, referring to application scenarios where the subjects present themselves to a fixed camera with a uniform background.

As can be observed from Table \ref{tab:DB}, the more recent face datasets have shifted towards the `in the wild' scenarios to consider more challenging variations. Nevertheless, since the two previously available LF face datasets were both collected in constrained conditions, it is not possible to assess the performance of LF-based face recognition solutions for unconstrained setups. The proposed LFFW dataset addresses this shortcoming by including 1908 LF images, corresponding to 429,300 2D images, from 53 subjects, thus providing a large-scale publicly available LF face dataset captured in the wild. Since this dataset will be publicly available, it may be used as the basis for the future design, validation, and assessment of LF-based face recognition systems. The proposed LFFC dataset complements the LFFW dataset by including 1060 LF and 238,500 facial 2D images from the same 53 subjects, captured between 1 day and 3 years before the corresponding LFFW acquisition. {It must be highlighted that while the LFFC acquisition conditions are similar to the previously available IST-EURECOM LFFD dataset \cite{iwbf}, the subjects are partly different.} The newly proposed LFFW and LFFC datasets offer from now on a unique publicly available data platform as this dataset combination provides precious information for the face recognition community, for instance to study aging effects and cross-dataset generalization performance of different analysis solutions.

\section{CapsField Solution}
This section presents the proposed face and expression recognition solution, exploiting the spatio-angular information available in LF images by using a combination of two deep learning networks.

\subsection{Problem Formulation}
Given a dataset $\mathcal{D}$ of LF face images, where each LF image $LF_{i, j}$ has been captured from a known face/expression with label $x_i \in \{x_1,x_2,\dots, x_N\}$ and a set of viewing angles $\theta_j \in \{\theta_{1}, \theta_{2}, \dots, \theta_{V}\}$, the tasks of face or expression recognition for a probe face sequence $LF_{probe}$ can be defined as:
\begin{equation} \label{eq:task}
    \hat{x} = arg \max\limits_{x_i} Pr(x_i|LF_{probe}, \mathcal{D}),
\end{equation}
where $Pr(x_i|LF_{probe}, \mathcal{D})$ is the probability of LF image $LF_{probe}$ belonging to label $x_i$.

\subsection{Solution Intuition and Overview}
 An LF image offers intra-view/spatial (within each view) and inter-view/angular (across views) information that can be used for improving face and expression recognition performance. Spatial features can be learned from each view using powerful CNN architectures. The inter-view information can be explored by combining the spatial features of individual views, for instance using feature level fusion, and feeding the result to a fully connected layer and then a classifier. However, the fully connected layer merely applies a weight vector to the concatenated feature vector, applying a non-linear function to the scalar output of a linear filter. This approach considers the concatenated features as a whole, not allowing to learn the relative positions of relevant face components in each view or across the different viewpoints, which is the main problem addressed in this paper.

A capsule network includes layers of capsules, each consisting of a group of neurons representing different properties of the input. The mechanism to learn interactions between capsules is performed with routing algorithms such as dynamic routing \cite{capsule}. Capsules can share the knowledge across locations using a pose matrix that represents the relation between parts of an object and the whole object, thus making capsules capable of dealing with viewpoint changes. Accordingly, the activation of a capsule is also based on a comparison between multiple incoming pose predictions. Capsule networks have recently been adopted in the context of face analysis tasks, most notably in age and gender classification \cite{CapFace1}, and action detection \cite{CapFace2}. This paper proposes the combination of a capsule network with a CNN to, respectively, learn the available inter-view and intra-view relations available in an LF image for face and expression recognition. In this context, the novel CapsField solution not only learns viewpoint-invariant relations between different parts of a face and the whole face in each view, but also it learns the relations between the different views along the whole multi-view sequence. 
A capsule network is also capable of learning feature importance by assigning higher weights to the more relevant features, while ignoring the spurious dimensions \cite{Patrick}. This functionality can act as an attention mechanism in the proposed CapsField solution, selectively focusing on the most important angular features making the solution more robust to changes in appearance and environment.

\begin{figure*}[!t]
\centering
\includegraphics[width=.96\linewidth]{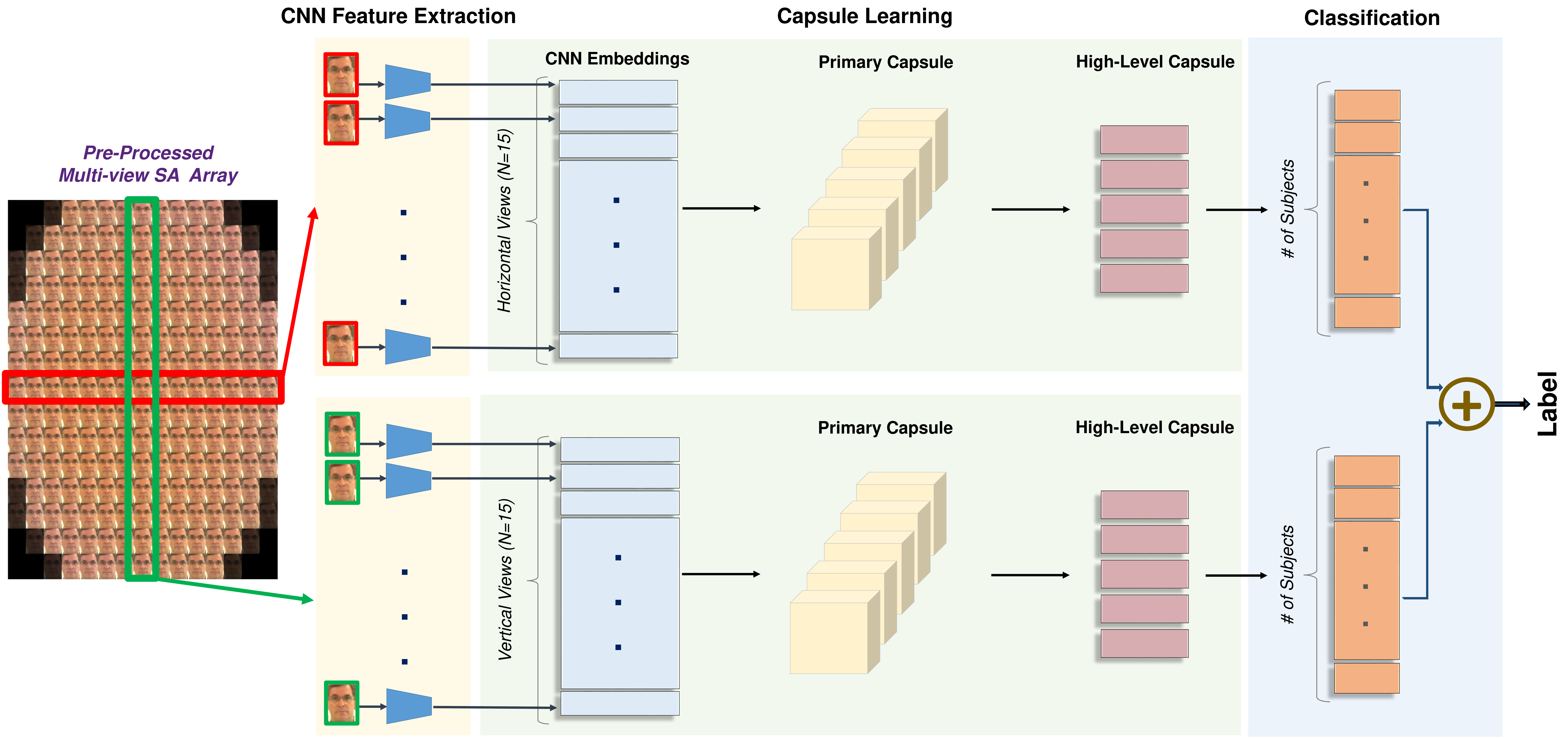}
\caption{Architecture of the CapsField solution.}
\label{fig:arch}
\end{figure*}

Following the motivation and intuition described above, the proposed CapsField solution integrates three sub-networks that are applied to the horizontal and vertical views to predict a classification lable, $L$, as follows:
\begin{equation}\label{eq:model_preview}
    L = (f_{CNN}^{H}\circ f_{CAPS}^{H}\circ f_{CLS}^{H}) + (f_{CNN}^{V}\circ f_{CAPS}^{V}\circ f_{CLS}^{V}),
\end{equation}
where $f_{CNN}^{H}$ and $f_{CNN}^{V}$ are CNN feature extraction sub-networks, respectively, applied to horizontal and vertical spatial SA sequences; these sub-networks are composed of multiple convolutional, pooling, and fully connected layers. $f_{CAPS}^{H}$ and $f_{CAPS}^{V}$ are capsule routing sub-networks that learn deeper part-whole relations between the spatial features and then selectively assign higher weights to the more discriminative features while ignoring misleading ones. The last pair of sub-networks, $f_{CLS}^{H}$ and $f_{CLS}^{V}$, respectively, perform classification for horizontal and vertical learned features using a softmax activation function. The results of the these three sub-networks, independently applied to horizontal and vertical views, are then fused using a sum-rule strategy, thus obtaining the predicted label, $l$. 

\subsection{Network Architecture and Walkthrough} 
Figure \ref{fig:arch} represents the architecture of the proposed CapsField solution composed by a pre-processing step and three sub-networks: CNN, capsule, and classification, whose descriptions are provided in the following.

\subsubsection{Pre-processing}
This paper uses the Light Field Toolbox v0.4 software \cite{LFTool} to render the lenslet Light Field Raw (LFR) images, thus creating a multi-view SA array with 15$\times$15 SA images, as illustrated in Figure \ref{fig:arch}-left. 
At this stage, the face region is cropped within each 2D SA image using the provided face bounding box locations available in the datasets. Next, the middle row and the middle column SA images from the multi-view SA array are scanned into two separate SA image viewpoint sequences (as shown by red and green rectangles in Figure \ref{fig:arch}-left), thus capturing the viewpoint changes along the horizontal and vertical directions, respectively. These two separate SA image viewpoint sequences are used as inputs to the proposed deep learning network.

\subsubsection{CNN Feature Extraction}
The proposed CapsField solution first utilizes CNNs to independently extract spatial features from the two input sequences, i.e., horizontal and vertical SA sequences. To this end, the VGG-16 \cite{pvz15} and Resnet50 \cite{resnet} CNN architectures have been, respectively, used for the face and expression recognition tasks. As both CNNs have been trained on the large-scale VGG-Face 2 dataset \cite{csxpz18}, the proposed solution will not suffer from overfitting and there is no additional training performed at this stage. More details about the used pre-trained models are provided in Section V-D.

\subsubsection{Capsule Routing}
A capsule network includes layers of capsules where each capsule consists of a group of neurons representing the information being learned. Capsule networks \cite{capsule} are composed of primary and secondary layers.
The first layer encapsulates the input using convolutional, reshaping, and squashing operations to represent the instantiating parameters, such as positional attributes, in the form of a vector. The secondary layer then learns deeper part-whole relations between the sub-parts of an object and the whole object.
In the CapsField solution proposed in this paper, the convolution operation has been removed from the primary capsule layer, so that only reshaping and squashing functions are applied to the spatial features coming from the previous sub-network. This modification is due to the fact that the previous CNN feature extraction sub-network already encodes the spatial information through several convolutional layers.

The output generated by the CNN feature extraction sub-network, i.e., the sequence of multi-view features, is first reshaped to a $ N_{c} \times C_{s}$ tensor where $N_{c}$ and $C_s$ are the number of primary capsules and their size, respectively. This operation is followed by a non-linear squashing activating function, creating an output vector in which each vector element takes a value in the [0,1] range:
\begin{equation}\label{eq:squash_function}
    \mathbf{v}_{j}=\frac{\left\|\mathbf{s}_{j}\right\|^{2}}{1+\left\|\mathbf{s}_{j}\right\|^{2}} \frac{\mathbf{s}_{j}}{\left\|\mathbf{s}_{j}\right\|},
\end{equation}
where $v_j$ is the $j^{th}$ vector output of the primary capsule layer and $s_j$ is the $j^{th}$ capsule input formed by reshaping the output of the CNN feature extraction sub-network. The orientation of the output vector, ${v}_{j}$, represents the positional properties of the input. This means that when the viewpoint of a section of a face changes, the orientation of the vector also changes while its length remains unchanged.

The secondary capsule layer is constructed by $N_{c}$ capsules, where each capsule is updated by a dynamic routing process as originally proposed in \cite{capsule}. The iterative dynamic routing process first learns the coupling coefficients, $C_{i,j}$, between two capsules, $i$ and $j$, available in two subsequent layers. After the degree of agreement between them is computed using the routing softmax function:
\begin{equation}\label{eq:routing_function}
\mathbf{C}_{i,j}=\frac{\exp ({b}_{i,j})}{\sum_{k} \exp ({b}_{i,k})},\
\end{equation}
where ${b}_{i,j}$ is the log probability indicating whether capsules $i$ and $j$ should be coupled.

Next, the input to capsule $j$, $s_{j}$, is computed as follows:
\begin{equation}\label{eq:capsul_output}
    \mathbf{s}_{j}=\sum_{i} C_{i j} \mathbf{W}_{i j} \mathbf{v}_{i},
\end{equation}
where $\mathbf{W}_{i j}$ is a trainable pose matrix that encodes the relation between the view and the views' sequence, thus establishing an intrinsic capacity for learning angular information. Eq. (\ref{eq:capsul_output}) can also be considered as an attention layer, where the coupling coefficients play the role of attention weights. 

Finally, a squashing function (Eq. (\ref{eq:squash_function})) is used again to create the output vector of capsule $j$. The iterative dynamic routing process is repeated between the primary and secondary layers for $N_{r}$ iterations, and the final outputs are used to feed the classifier. 


\subsubsection{Classification}
The proposed network uses two independent dense layers with softmax activation as the classifier for the horizontal and vertical capsule features, thus computing the corresponding classification scores. Score-level fusion is finally applied, averaging the horizontal and vertical classification scores to compute the final class probability vector, determining the face or expression recognition result.

\subsection{Implementation and Training Details}\label{sec:implementation}
The parameter values for training the proposed network are empirically tuned to achieve the best performance for both the face and expression recognition tasks, as presented in Table \ref{tab:hyper}. This table presents the optimal parameter values for each sub-network as well as for the whole deep network, CapsField. As can be observed, the network settings for both face and expression recognition tasks are rather similar except for: i) the choice of CNNs, as the two tasks use different CNN models for spatial feature extraction; and ii) the training batch size, as the number of input samples is different for the face and expression recognition tasks. {Concerning the number of routing iterations, different values ranging from 2 to 6 have been considered, where increasing the value to more than 3 did not improve the performance while increasing the computational time.} The entire architecture has been implemented using TensorFlow \cite{tensorflow} with Keras backend and is trained using a GeForce GTX 1080 GPU.

\begin{table}
  \centering
  \setlength\tabcolsep{3pt}
\footnotesize
    \caption{Network parameters setting.}
    \begin{tabular}{l|l|l|l}
    \hline
    \textbf {Layer}& \textbf{Parameter} &\textbf{Face rec.}& \textbf{Expression rec.}\\
    \hline\hline
     CNN & CNN type& ResNet50 & VGG-16 \\
     feature& Pre-trained model& VGG-Face2 & VGG-Face2 \\
     extraction& Embeding layer & Average pooling & FC6  \\
     & Embeding size & 2048 & 4096 \\
     \hline
     Capsule & \# of capsules & 5 & 5 \\
     routing& Capsule size & 64 & 64 \\
     & \# of routing itetrations & 3 & 3 \\
     \hline
     CapsField  & Mini-batch siz & 53 & 52 \\
     network & Loss function & Cross-entropy & Cross-entropy \\
     & Optimizer & rmsprop & rmsprop \\
     & Metric & Accuracy & Accuracy \\
    \hline
    \end{tabular}%
  \label{tab:hyper}%
\end{table}%

\section{A Novel Light Field Faces in the Wild Dataset}
The proposed LFFW dataset fills the gap regarding the lack of availability of a LF face dataset captured in the wild. It includes data from 53 subjects, 42 males and 11 females, with 36 LF image shots per person, covering several unconstrained variations, totaling 1908 LF face images in the dataset. As each LF image consists of 225 2D SA images captured from 15$\times$15 different perspectives, the LFFW dataset includes 429,300 (53$\times$36$\times$225) 2D facial images in total. The participants were between 20 and 62 years old, from 10 different countries across the globe. The unconstrained variations considered in the proposed LFFW dataset can be categorized into two types: i) acquisition variations, notably different dates, environments, distances, and lighting conditions; and ii) facial variations, notably facial poses, random expressions, occlusions, and actions {with no pre-defined protocols or any stimuli used to induce the desired variations}. All LF images were taken under ‘in the wild’ conditions, meaning that no restrictions were imposed on the subjects when capturing their photos. The LF images have been captured using a Lytro ILLUM LF camera \cite{Lytro}, with a 40 Megaray sensor and a 30-250 mm lens with 8.3x optical zoom and f/2.0 aperture. 

In addition to the LFFW dataset, this paper also proposes a complementary, constrained dataset with the same 53 LFFW subjects, named LFFC, which includes their facial images captured in a controlled acquisition setup. The LFFC LF images have been acquired between 1 day and 3 years prior to acquiring the corresponding in the wild images. The LFFC dataset includes 20 LF shots per person, with different facial variations including expressions, actions, poses, illuminations and occlusions, thus providing 238,500 (53$\times$20$\times$225) 2D facial images in total. This dataset provides additional information essential for studying aging effects and the constrained versus in the wild cross-dataset generalization ability.

\subsection{LFFW Dataset Structure}
\begin{figure}[!t]
\centering
\includegraphics[width=0.86\columnwidth]{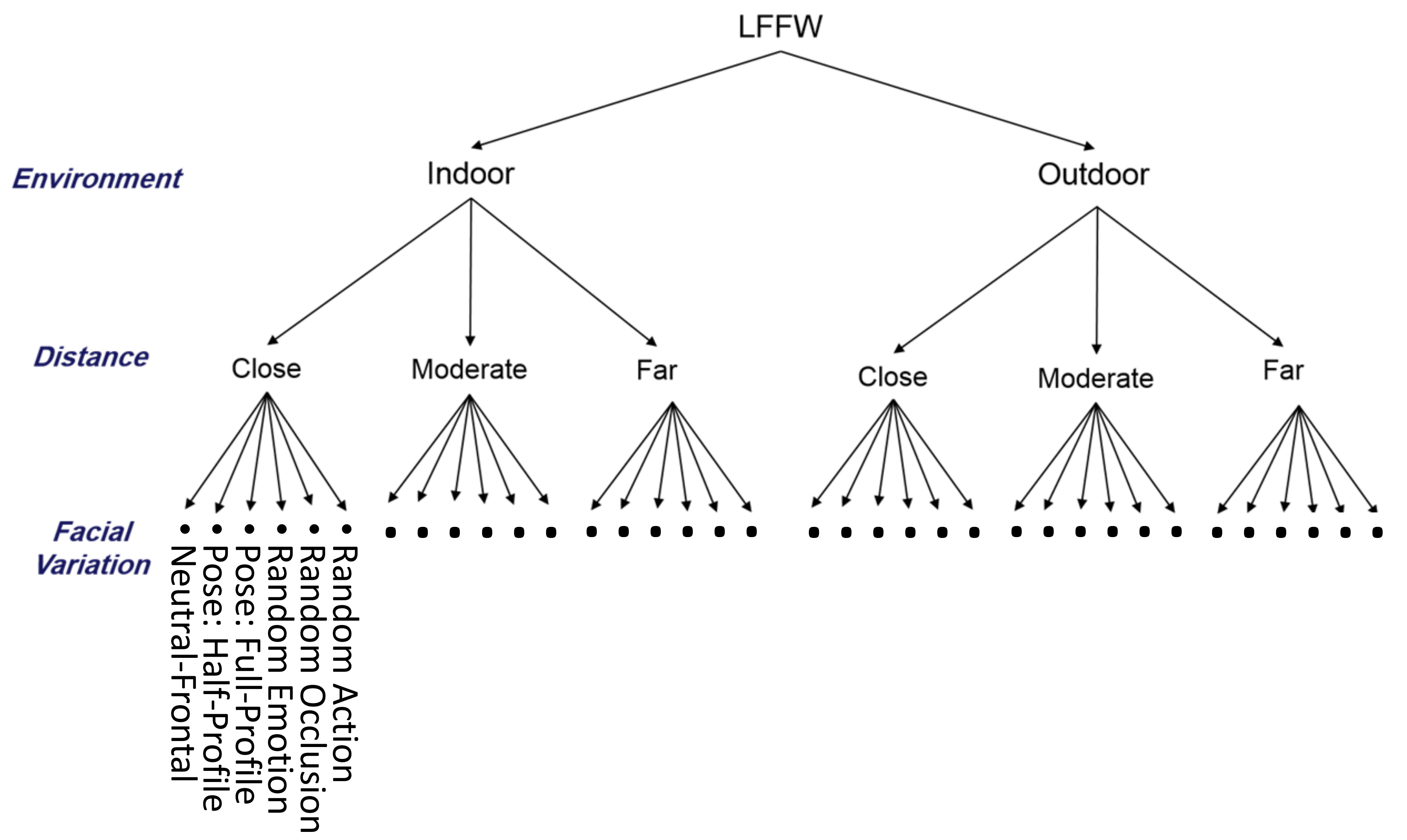}
\caption{Hierarchical structure of the proposed LFFW dataset.}
\label{fig:structure}
\end{figure}

\begin{figure}[!t]
\centering
\includegraphics[width=0.85\columnwidth]{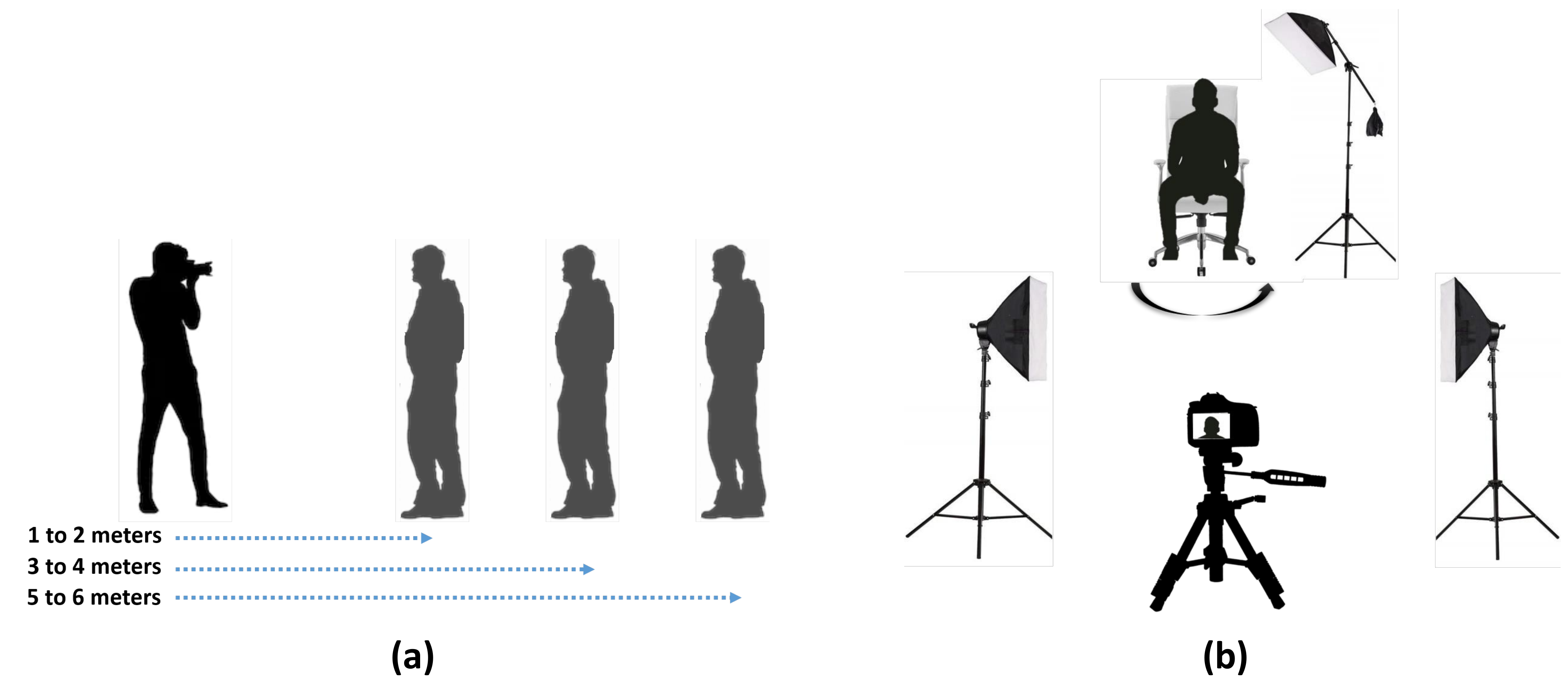}
\caption{{Illustration of acquisition setup for (a) LFFW and (b) LFFC .}}
\label{fig:sketch}
\end{figure}

\begin{figure}[!t]
\centering
\includegraphics[width=1\columnwidth]{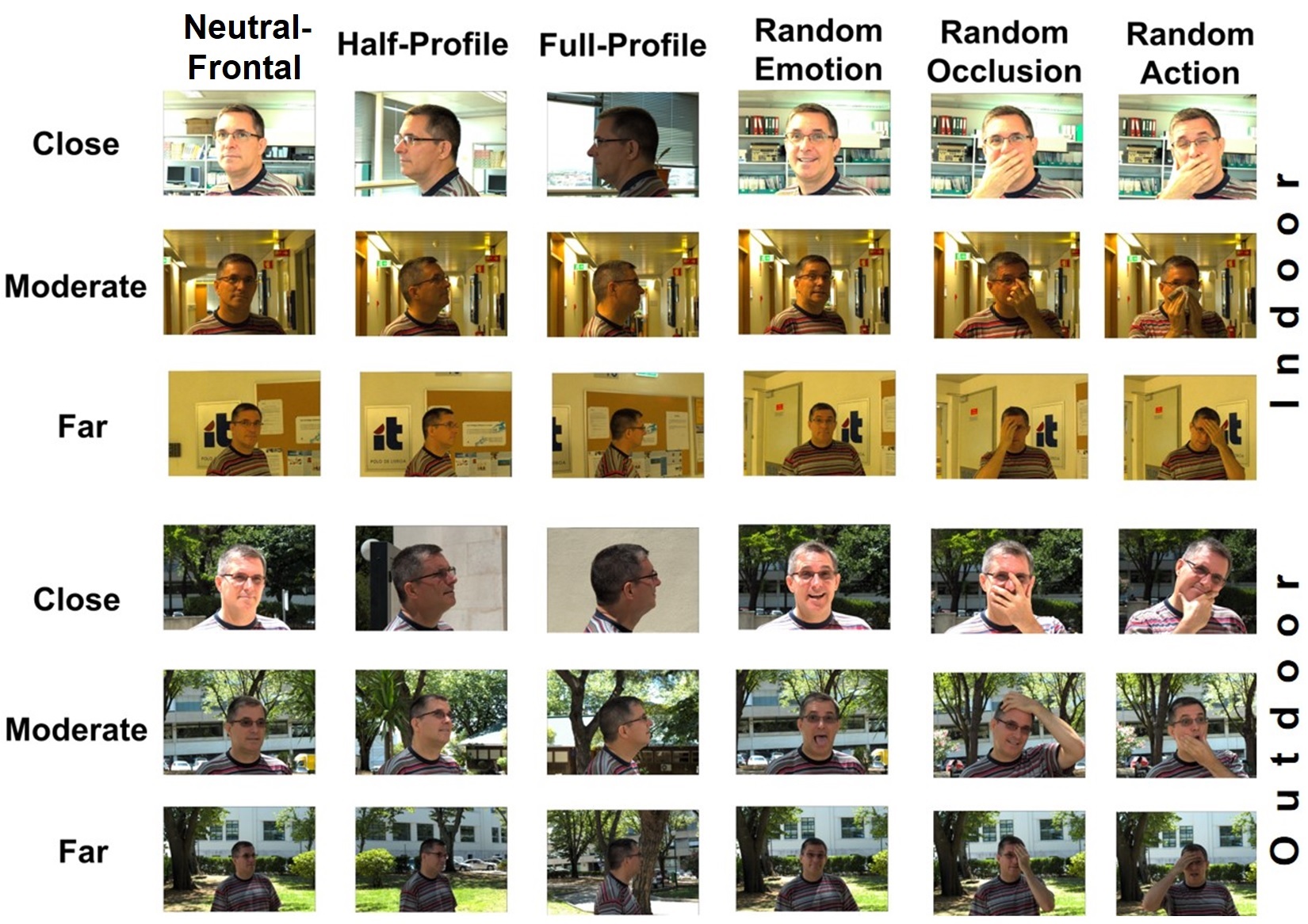}
\caption{Illustration of the 2D central SA images for the variations of a specific subject in the proposed LFFW dataset.}
\label{fig:LFFW}
\end{figure}

The hierarchical structure designed for the LFFW dataset is illustrated in Figure \ref{fig:structure}. The LF image acquisition has been performed in both indoor and outdoor environments, at different locations, with different backgrounds. The acquisition has also been performed in different lighting conditions, e.g., from gloomy to extremely sunny weather, allowing images to be captured in the wild with different illuminations and in the presence of uncontrolled shadows, and backgrounds. {As shown in Figure \ref{fig:sketch}(a),} for each environment, LF images were captured from three distance ranges: close, moderate, and far, respectively captured from around 1 to 2, 3 to 4, and 5 to 6 meters from the subjects, thus providing facial LF images with different face spatial resolutions. Finally, the LF images have been captured with several facial variations, including different poses, notably frontal, half-profile, and full-profile, random expressions, occlusions, and actions. {This process has been repeated for each environment-distance pair, which means each subject acted each facial variation 6 times.} Examples of the variations considered for a subject in the LFFW dataset are illustrated in Figure \ref{fig:LFFW}.

\subsection{LFFW Dataset Elements}\label{sec:problem_definition}
The LFFW dataset is composed by the following elements: 
\begin{enumerate}
\item \textbf{Raw LF Images:} LF face images are stored in raw file format, so-called LF Raw (LFR) files. The average size of each file is approximately 50 MB. LFR files can be used as input for the Lytro Desktop Software \cite{LytroDesk} or any other processing library/toolbox, such as the Matlab Light Field Toolbox V0.4 \cite{LFTool}.
\item \textbf{2D Central SA Images:} The proposed dataset also includes the 2D SA images for the central view of each LF image, generated using the Matlab Light Field Toolbox V0.4 \cite{LFTool}. Naturally, other 2D SA images corresponding to different perspectives can be directly rendered from the available raw LF images. 
\item \textbf{Face Bounding Boxes Information}: The face bounding box locations for the 2D central SA images, detected using the YOLO face detector \cite{rdgf16}, are included in the dataset. 
\item \textbf{Subjects Metadata Information:} The LFFW dataset includes rich metadata regarding the acquisition date, as well as information on the subjects' gender and age.
\item \textbf{Calibration Information:} Calibration data, which is critical for compensating the specific properties of the used LF camera, is an essential component in the dataset. For example, this is a required input for the Lytro Desktop Software \cite{LytroDesk} and the Matlab Light Field Toolbox \cite{LFTool}.
\end{enumerate}

\subsection{LFFC Dataset}
This paper also proposes an LF Face Constrained (LFFC) dataset, complementary to the LFFW dataset; it includes 1060 LF or 238,500 2D SA facial images captured in a controlled environment, from the same 53 subjects available in LFFW. This pair of complementary datasets has considerable research potential. Image acquisition was performed in an indoor environment, using the same Lytro ILLUM lenslet camera \cite{Lytro} similar to the LFFW dataset. {An illustration of the LFFC acquisition setup is presented in Figure \ref{fig:sketch}(b);} the acquisition setup included a white backdrop background behind a chair for the subject, placed at a fixed distance of 1.25 m from the camera. The scene was illuminated with a three-point lighting kit, including key, fill, and back lights, placed to limit shadows and allow easy segmentation of the subject from the background.

\begin{figure}[!t]
\centering
\includegraphics[width=0.90\columnwidth]{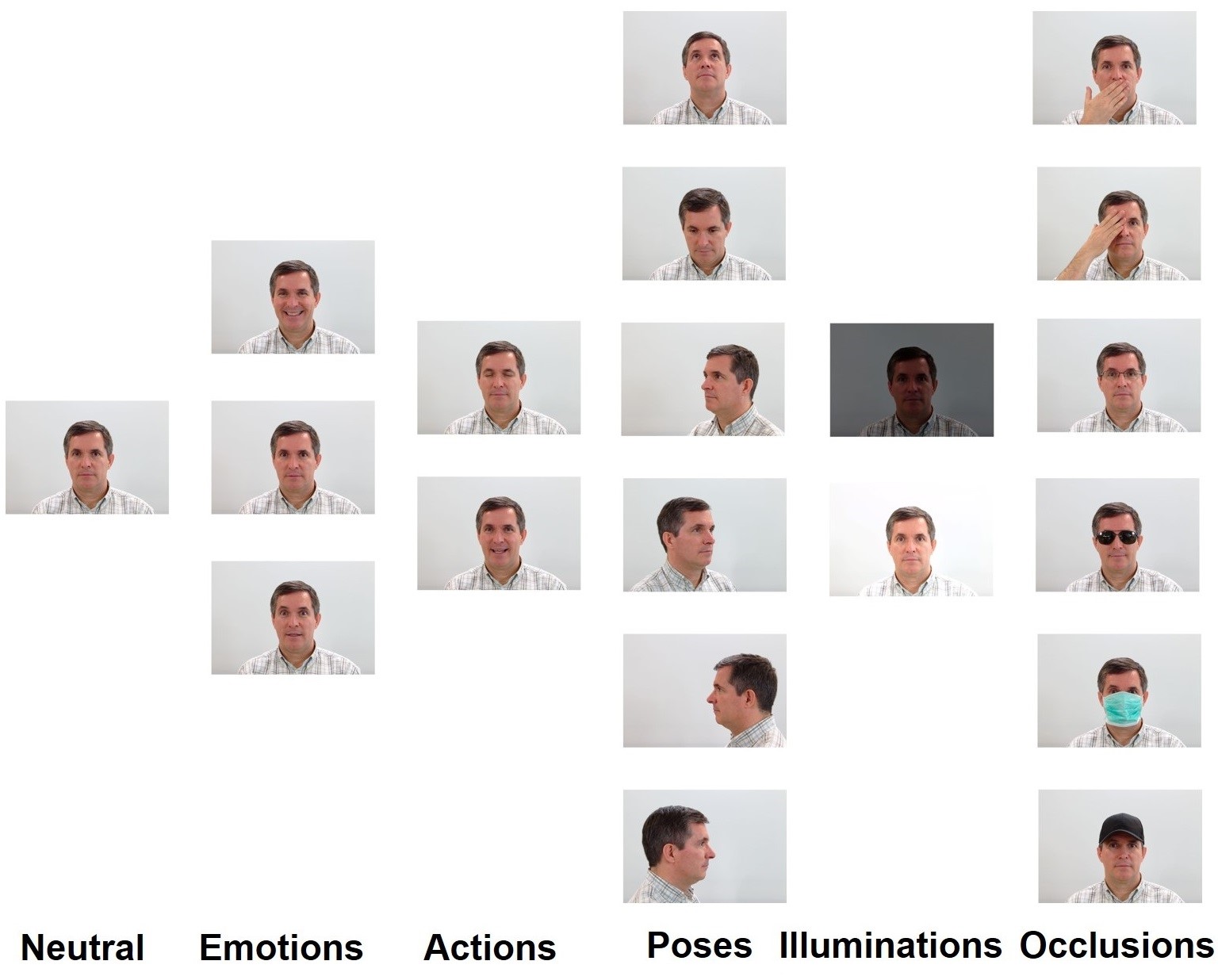}
\caption{Illustration of the 2D central SA images for the variations of a specific subject in the proposed LFFC dataset.}
\label{fig:LFFC}
\end{figure}

The LFFC dataset includes 20 LF shots per person (each including 225 2D SA images), with different facial variations including: i) one neutral image; ii) three facial expressions: happiness, anger and surprise; iii) two facial actions: closed eyes and open mouth; iv) six poses: looking up, looking down, right half-profile, right profile, left half-profile, left profile; v) two different illuminations: low and high illumination levels; and vi) six facial occlusions: eye and mouth occluded by hand, and face occluded by glasses, surgical mask, and hat. Examples of the various face variations considered in the LFFC dataset are illustrated in Figure \ref{fig:LFFC}.

\begin{figure}[!t]
\centering
\includegraphics[width=0.6\columnwidth]{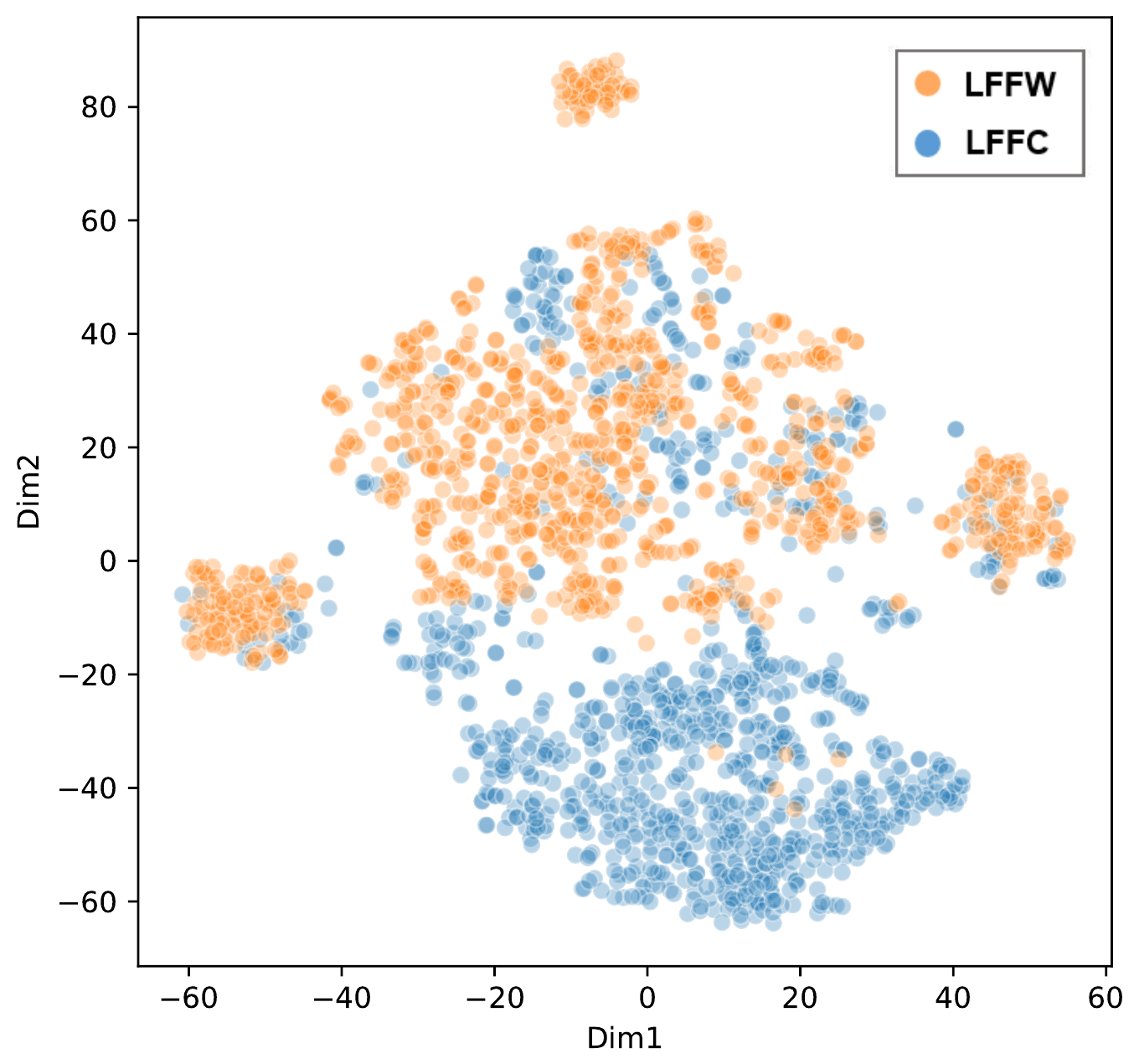}
\caption{Visualization of LFFW and LFFC distributions using t-SNE.}
\label{fig:tsne_dist}
\end{figure}

The availability of the LFFC dataset enables constrained versus in the wild cross-dataset studies.
To the best the authors' knowledge, this type of study has not been yet performed for LF-based recognition due to the difficulties in obtaining the required data. Additionally, the temporal gap between the acquisition of the two datasets also allows  to study the impact of aging. 
In order to illustrate how challenging cross-dataset studies can be, t-Distributed Stochastic Neighbor Embedding (t-SNE) visualization \cite{tSNE} has been used to summarize the distribution of images available in the two LFFW and LFFC datasets in a two-dimensional space (see Figure \ref{fig:tsne_dist}). t-SNE maps the multi-dimensional input data to a lower dimensional feature space.
The figure shows that considerable difference exists between the two datasets, further indicating that generalization across the two datasets (training on one and testing on the other) may be difficult.


\subsection{{Ethics, Access, and Usage Conditions}}

{The participants in the capture of both datasets all signed a consent form before the acquisition, allowing the collection of LF images of their faces, to become part of the proposed datasets, naturally, to be used for research purposes. All access to the dataset images is controlled and supervised, and images will be used anonymously. The generic information asked from the subjects (such as age or gender), to be included as metadata, will also be made available anonymously. Only images from some specific subjects (shown in Figure \ref{fig:exp}) may appear in publications as per received consent.}


\subsection{Facial Expression Annotation}
\begin{figure}[!t]
\centering
\includegraphics[width=1\columnwidth]{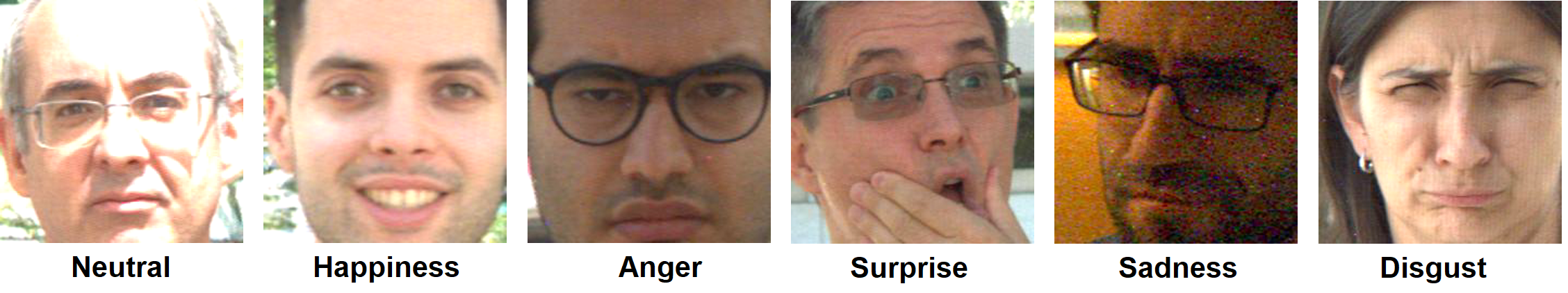}
\caption{Illustration of cropped 2D SA images with different facial expressions from the LFFW dataset.}
\label{fig:exp}
\end{figure}

The LFFW dataset includes facial images that are captured with random expressions. To support their usage for expression recognition algorithms, {the random expressions are carefully analyzed by the authors of this paper in order to validate and label the acted expressions}. In this context, 832 face LF images have been manually annotated with different expression labels including: neutral, happiness, anger, surprise, sadness and disgust, as illustrated in Figure \ref{fig:exp}. As can be seen, some face expressions are captured in very challenging acquisition conditions, e.g., at very high or very low illumination levels or with occlusions by hand or glasses, making the expression recognition task more difficult. 

\section{Performance Assessment}
This section presents the proposed experimental evaluation protocols and the corresponding performance results and analysis for the proposed and benchmarking recognition solutions. This section also performs ablation experiments, investigating the effect of the individual sub-networks on the overall recognition performance.

\subsection{Face Recognition Evaluation Protocols}
To assess the overall face recognition performance for in the wild conditions, five evaluation protocols using the LFFW and LFFC datasets are proposed. {These protocols divide the whole data into training and testing sets as described below, where 10\% of the training set images have been used for validation in all the test protocols.}

\subsubsection{Cross-environment Evaluation Protocol} {This protocol uses the 18 indoor LFFW LF images of each subject for training, whereas the testing phase uses the LFFW outdoor LF images. Then, the process is repeated using the outdoor images as training data and the indoor images as test data.} For this protocol, performance results are reported with respect to the various facial variations. Moreover, the sensitivity of the face recognition solutions to the unconstrained environmental conditions in the wild, such as lighting, shadows, occlusions, and backgrounds, is analyzed.

\subsubsection{Cross-distance Evaluation Protocol} {In this protocol, the training stage uses the 12 LFFW LF images of each subject captured from one of the three distances (close, moderate and far), whereas the testing phase independently uses the 24 captured LF images from the two other distances. This process is repeated, changing the training and testing (distance) sets, for a total of six different iterations.} This protocol evaluates the face recognition solutions’ sensitivity to various details in the face region resulting from capturing images at different distances and in the wild.

\subsubsection{Cross-pose and expression Evaluation Protocol} {This protocol considers practical scenarios where only frontal-neutral LF images are available in the training dataset, and LF facial images captured in the wild are used for testing.} Therefore, the training set contains only the 6 frontal-neutral LFFW LF images of each subject, while the testing set considers all the other facial variations. 
    
\subsubsection{Cross-dataset Evaluation Protocol} {This protocol performs the training with the LFFC dataset and the testing with the LFFW dataset, and vice-versa.} The performance results are reported with respect to the different facial variations. This protocol evaluates the generalization ability of the face recognition solutions, and to some extent the robustness to aging effects. This type of assessment is rarely found in the literature as it requires the availability of appropriate complementary datasets as available in this work. This type of assessment is performed here, for the first time, with LF images.

\subsubsection{{Subject-Independent Evaluation Protocol}} {This protocol performs the training with all the images from the first 28 subjects (ID No. 1 to 28) in the LFFW dataset. The trained model is then used as a feature extractor for the testing set including all the images from the remaining 25 subjects. Then, in the testing set, the frontal images have been used as the gallery samples and the rest of the images have been used as probe samples to perform the recognition process. In this protocol, the testing identities are disjoint from the training ones, thus making the face recognition process more challenging and closer to real-world application scenarios.}

\subsection{Expression Recognition Evaluation Protocols}
To assess the performance of the face expression recognition solutions using the annotated face images, two evaluation protocols are proposed. It should be noted that the number of evaluation protocols designed for expression recognition is less than for face recognition since the expression samples are comparatively limited, both in terms of numbers and variations.

\subsubsection{{Subject-Independent} 4-fold Cross Validation} The first evaluation protocol performs {subject-independent} 4-fold Cross Validation (CV) on the LFFW dataset. Accordingly, the proposed expression recognition solutions are trained with 1/4 of the annotated images available in each expression class, while the remaining 3/4 of the LF images are used for testing. This process is repeated 3 more times with the next folds and the average results are reported.

\subsubsection{Cross-dataset Evaluation Protocol} This protocol considers cross-dataset evaluation by using expression samples from the LFFC dataset for training and the LFFW dataset for testing, and vice-versa. The in the wild and constrained samples were, respectively, categorized into 6 and 4 expression classes. The two additional expression classes for the in the wild dataset not available in the constrained dataset, i.e., sadness and disgust, are naturally removed from the experiments for this evaluation protocol.

\subsection{Benchmarking Recognition Solutions}
Since this paper proposes new evaluation protocols for the two developed datasets, the authors have re-implemented all the benchmarking solutions for both face and expression recognition tasks since otherwise no comparative results would be available. To do so, the best configuration of parameters reported in the original papers have been used to report the rank-1 performance results in this paper.
 
\subsubsection{Face Recognition} The LF-based benchmarking solutions used for the face recognition task include LFLBP \cite{icip}, LFHG \cite{ear}, VGG-D3 \cite{MLSP}, ResNet50+ST-LSTM \cite{LSTMV1}, VGG+Conv-LSTM \cite{CSVT}, VGG+GLF-LSTM \cite{joint}, VGG+SLF-LSTM \cite{joint}, and ResNet50+DS-LSTM \cite{LSTMV2} that have been introduced in Section III. Three non-LF solutions including the VGG-16 \cite{pvz15}, Resnet50 \cite{csxpz18}, SE-ResNet-50 \cite{csxpz18}, and ArcFace with ResNet-101 backbone \cite{arc} networks, have also been used in the comparisons. Naturally, these non-LF solutions only process the 2D central SA image. {It must be noted that some of LF-based face recognition solutions, e.g., \cite{ryrb13, rryb13, rrb16}, consider the \textit{a posteriori} refocusing capability of LF cameras to improve the image quality of the already captured face images for 2D face recognition. Naturally, these methods are not comparable with the proposed solution which deals directly with the LF data, in its native 4D multi-view data format, to exploit the available spatio-angular information.}

\subsubsection{Expression Recognition} Four available LF-based solutions including ResNet50+ST-LSTM \cite{LSTMV1}, VGG+LSTM \cite{ACII}, VGG+BiLSTM+Attention \cite{ICASSP}, and ResNet50+DS-LSTM \cite{LSTMV2}, as introduced in Section III-C, are used. The benchmarking solutions also include the VGG-16 \cite{csxpz18} network, pre-trained with VGG-Face 2.

\subsection{Comparative Face Recognition Performance}
This section reports the rank-1 face recognition rates, for the four evaluation protocols proposed earlier, when using CapsField and 12 state-of-the-art face recognition solutions, with the best results highlighted in bold.


\subsubsection{Cross-environment Performance} 
The results in Table \ref{tab:env} refer to the cross-environment protocol and are independently reported for different facial variations, showing that the environment variations, including different lighting conditions, shadows, occlusions, and non-uniform backgrounds can negatively impact the face recognition performance in the wild. The proposed CapsField solution delivers very impressive performance results when trained with indoor and tested with outdoor samples. The proposed solution achieves less impressive results when tested with indoor samples, showing that it is sensitive, notably to low lighting conditions. Overall, the results show that, on average, the proposed CapsField solution is superior to the other benchmarking solutions for this evaluation protocol. 

\subsubsection{Cross-distance Performance} 
The cross-distance protocol results are presented in Table \ref{tab:dist}. Since the Lytro ILLUM LF camera used in this research has a small baseline, it is expected that more disparity is captured at closer distances.  {Additionally, the cropped face images have different resolutions, due to the different capturing distances. As the CNN feature extractor first resizes all the images to $224 \times 224$ pixels, the images with larger or smaller resolutions were, respectively, down-sampled or up-sampled to be then processed by the network; this means the images captured at far distances may include up-sampling artefacts.}
Hence, the overall performance results for all benchmarking solutions are better when testing with the images captured from close and moderate distances, as these LF images feed the recognition solutions with more details, thus further contributing to an improved performance. The proposed CapsField solution obtains very good results, notably achieving a performance gain of 2.95\% when compared to the best performing benchmarking solution, i.e., VGG+SeqL-LSTM \cite{joint}.

\begin{table*}[]
\centering
\setlength\tabcolsep{2.5pt}
\footnotesize
\centering
\caption{Cross-environment protocol assessment: Face recognition rank-1 results.}
\begin{tabular}{l|l|llllll|llllll|l}
\hline

 \multicolumn{1}{ c |}{\textbf{Solution}}& \textbf{Year} & \multicolumn{6}{ c }{\textbf{Train: \textit{Indoor}, Test: \textit{Outdoor}}} & \multicolumn{6}{ |c }{\textbf{Train: \textit{Outdoor}, Test: \textit{Indoor}}} & \multicolumn{1}{ |c }{\textbf{Avg.}}  \\ 
 
\hline 
         & & \textbf{{\vtop{\hbox{\strut Neutral}\hbox{\strut Frontal}}}} & \textbf{{\vtop{\hbox{\strut Rand.}\hbox{\strut Exp.}}}}   & \textbf{{\vtop{\hbox{\strut Half-}\hbox{\strut Profile}}}}   & \textbf{{\vtop{\hbox{\strut Full-}\hbox{\strut Profile}}}} & \textbf{{\vtop{\hbox{\strut Rand.}\hbox{\strut Occlus.}}}}  & \textbf{{\vtop{\hbox{\strut Rand.}\hbox{\strut Action}}}} & \textbf{{\vtop{\hbox{\strut Neutral}\hbox{\strut Frontal}}}} & \textbf{{\vtop{\hbox{\strut Rand.}\hbox{\strut Exp.}}}}   & \textbf{{\vtop{\hbox{\strut Half-}\hbox{\strut Profile}}}}   & \textbf{{\vtop{\hbox{\strut Full-}\hbox{\strut Profile}}}} & \textbf{{\vtop{\hbox{\strut Rand.}\hbox{\strut Occlus.}}}}  & \textbf{{\vtop{\hbox{\strut Rand.}\hbox{\strut Action}}}} &   \\

\hline\hline

VGG-16 \cite{pvz15}&   2015     & 96.23\% & 94.33\%         & 91.19\%         & 38.36\%      & 86.79\%      & 72.96\%       & 91.19\%    & 86.16\% & 69.18\%         & 18.24\%      & 67.30\%      & 55.35\%       & 72.27\%       \\

LFLBP \cite{icip}&      2017     & 10.06\% & 11.32\%         & 11.95\%         & 11.32\%      & 9.43\%       & 7.55\%        & 8.18\%     & 6.92\%  & 3.77\%          & 3.77\%       & 3.77\%       & 5.66\%        & 07.81\%        \\
SE-ResNet-50 \cite{csxpz18} &      2018     & 96.22\%  & 93.71\% & 91.82\% & 74.84\% & 81.76\% & 81.13\% & 86.16\% & 85.53\% & 79.87\% & 50.94\% & 68.55\% & 59.74\% &  79.61\%        \\
ResNet-50 \cite{csxpz18} &      2018     & 97.48\%  & 98.11\% & 98.11\% & 83.01\% & 91.19\% & 88.67\% & 89.93\% & 85.53\% & 74.21\% & 50.31\% & 77.35\% & 66.66\% & 84.17\%        \\
LFHG \cite{ear}&      2018      & 23.90\% & 30.19\%         & 25.79\%         & 22.64\%      & 19.50\%      & 21.38\%       & 29.56\%    & 31.45\% & 19.50\%         & 22.01\%      & 20.13\%      & 20.13\%       & 23.85\%      \\
VGG-D3  \cite{MLSP}&    2018      & 95.60\% & 93.71\%         & 91.19\%         & 38.36\%      & 86.16\%      & 71.70\%       & 91.82\%    & 85.53\% & 67.30\%         & 16.99\%      & 65.41\%      & 54.72\%       & 71.54\%   \\
{ResNet50+ST-LSTM} \cite{LSTMV1}& 2018 & 98.13\% &  99.37\%  &  96.85\%  &  89.31\%  &  89.94\%    &  88.68\%  &  90.56\%  &  90.56\% &  76.72\%  &  55.97\%   &   79.25\%  & 71.07\%  &  85.53\%   \\
VGG+Conv-LSTM  \cite{CSVT} & 2019 & 98.74\% & 96.23\%         & 95.60\%         & 77.99\%      & 91.82\%      & 79.87\%       & 93.08\%    & 90.57\% & 81.76\%         & 68.55\%      & \textbf{84.28\%}      & 71.07\%       & 85.80\%    \\
{ArcFace} \cite{arc}& 2019 &  94.97\% &  96.23\%  &  91.19\%  &  77.35\%  &  92.45\%    &  91.19\%  &  88.05\%  &  89.94\% &  77.99\%  &  66.04\%   &   79.25\%  & 71.70\%  &  84.70\%   \\ 
VGG+GLF-LSTM  \cite{joint} & 2020  & 98.74\% & 96.23\%         & 95.60\%         & 77.99\%      & 91.22\%      & 81.13\%       & 94.34\%    & 90.57\% & 84.91\%         & 68.55\%      & 82.39\%      & 71.70\%       & 86.11\%    \\
VGG+SLF-LSTM \cite{joint}& 2020  & 98.11\% & 96.86\%         & 95.60\%         & 81.76\%      & 90.57\%      & 83.02\%       & \textbf{95.60\%}    & 89.94\% & \textbf{86.16\%}         & \textbf{70.44}\%      & 81.76\%      & 68.55\%       & 86.53\%    \\
{ResNet50+DS-LSTM} \cite{LSTMV2}& 2020 &  98.74\% &  98.74\%  &  96.85\%  &  86.79\%  &  92.45\%    &  88.67\%  &  86.16\%  &  86.79\% &  74.84\%  &  52.20\%   &   76.10\%  & 67.30\%  &  83.80\%   \\
Prop. ResNet50+Capsule & ---  & \textbf{100\%}  & \textbf{100\%}   & \textbf{99.37\%}   & \textbf{91.19\%}  & \textbf{94.33\%} & \textbf{91.82\%} & 90.56\% & \textbf{93.71\%} &  81.76\% & 55.97\% & 83.02\% & \textbf{72.33\%} & \textbf{87.83\%} \\ 
\hline

\end{tabular}
\label{tab:env}
\end{table*}

\begin{table*}[]
\centering
\setlength\tabcolsep{2.5pt}
\footnotesize
\centering
\caption{Cross-distance protocol assessment: Face recognition rank-1 results.}
\vspace{-3pt}
\begin{tabular}{l|l|llllll|l}
\hline

\textbf{Solution}   & \textbf{Year}                & \textbf{Train: Close}   & \textbf{Train: Close} & \textbf{Train: Moderate} & \textbf{Train: Moderate} & \textbf{Train: Far}  & \textbf{Train: Far}     & \textbf{Average} \\
                        & & \textbf{Test: Moderate} & \textbf{Test: Far}    & \textbf{Test: Close}     & \textbf{Test: Far}       & \textbf{Test: Close} & \textbf{Test: Moderate} &         \\
\hline\hline
VGG-Face  \cite{pvz15}       &2015 & 68.23\%        & 68.39\%      & 77.67\%         & 75.31\%         & 77.04\%     & 74.84\%        & 73.58\% \\
LFLBP  \cite{icip}          &2017& 13.83\%        & 11.01\%      & 16.03\%         & 15.56\%         & 12.89\%     & 16.66\%        & 16.26\% \\
SE-ResNet-50 \cite{csxpz18} &      2018     & 81.76\%  & 75.94\% & 90.25\% & 77.25\% & 83.30\% & 82.54\% & 81.84\%           \\
ResNet-50 \cite{csxpz18} &      2018     & 83.33\%  & 76.57\% & 95.44\% & 79.72\% & 92.45\% & 84.59\% &  85.35\%         \\
LFHG  \cite{ear}           &2018 & 25.47\%        & 23.42\%      & 26.57\%         & 24.52\%         & 24.52\%     & 25.00\%        & 24.92\% \\
VGG-D3  \cite{MLSP}          &2018 & 67.29\%        & 67.29\%      & 78.45\%         & 74.68\%         & 76.25\%     & 74.05\%        & 73.00\% \\
{ResNet50+ST-LSTM} \cite{ LSTMV1}& 2018 &  83.02\% &  73.27\%  &  96.86\%  &  80.82\%  &  95.75\%    &  89.15\%  &  86.47\%   \\
VGG + Conv-LSTM \cite{CSVT} &2019 & 83.78\%        & 82.16\%      & 85.63\%         & 85.76\%         & 88.28\%     & 87.10\%        & 85.45\% \\
{ArcFace} \cite{arc}& 2019 &  80.02\% &  72.74\%  &  79.08\%  &  77.04\%  &  78.45\%    &  76.57\%  &  77.31\%   \\
VGG + GLF-LSTM \cite{joint}  &2020 & 84.10\%        & 81.64\%      & 86.18\%         & 85.71\%         & 88.08\%     & 87.13\%        & 85.47\% \\
VGG + SLF-LSTM \cite{joint}  &2020 & 84.79\%        & \textbf{82.67}\%      & 85.70\%         & \textbf{86.33}\%         & 88.93\%     & 87.55\%        & 86.00\% \\
{ResNet50+DS-LSTM} \cite{LSTMV2}& 2020 &  82.08\% &  75.31\%  &  95.91\%  &  81.13\%  &  94.81\%    &  89.47\%  &  86.45\%   \\

\hline
Prop. ResNet50 + Capsule & ---        & \textbf{86.79}\%        & 79.24\%      & \textbf{97.80}\%         & 83.96\%         & \textbf{96.85}\%     & \textbf{89.93}\%        & \textbf{89.10}\%\\
\hline
\end{tabular}
\label{tab:dist}
\end{table*}

\subsubsection{Cross-pose and expression Performance} 
The cross-pose and expression protocol results are presented in Table \ref{tab:pose} for the several facial variations considered, notably including different face orientations, random expressions, occlusions, and actions. The overall performance results for the benchmarking solutions show poor performance when trained with only frontal faces and tested with different poses, notably full-profile images. The performance is also poor for random action variations as the type of actions during training can be very different from those used for testing. Overall, the performance of the proposed CapsField solution is considerably better than all other benchmarking solutions, notably showing a gain of 15.85\% over the best benchmark.

\subsubsection{Cross-dataset Performance}
The results in Tables \ref{tab:DB1} and \ref{tab:DB2} refer to the cross-dataset evaluation protocol and correspond to two situations: i) LFFC is used for training and LFFW is used for testing; and ii) LFFW is used for training and LFFC is used for testing. As expected, better performance results are obtained when testing on the less challenging LFFC dataset. The results show that the proposed solution outperforms all the benchmarking solutions for both cases by a large margin, achieving performance gains of 11.05\% (when testing with LFFC) and 14.62\% (when testing with LFFW), compared to the best performing benchmark. These results, showing a better generalization ability of the proposed CapsField solution, clearly illustrate the advantage of exploring the richer information available in LF images with the consideration of a more robust model created by capsule networks. 

\subsubsection{{Subject-independent Performance}} 
{The subject-independent protocol results are presented in Table \ref{tab:ind} for the facial variations considered, notably different poses, random expressions, occlusions, and actions. Overall, the average performance of the proposed CapsField solution is better than all the benchmarking solutions, even for this protocol, where the face recognition process is more challenging and closer to real-world application scenarios.}

\begin{table*}[]
\centering
\setlength\tabcolsep{4.5pt}
\footnotesize
\centering
\caption{Cross-pose and expression protocol assessment: Face recognition rank-1 results.}
\begin{tabular}{l|l|lllll|l}
\hline
\textbf{Solution}  & \textbf{Year} &      \textbf{Rand. Exp.} & \textbf{Half-Profile} & \textbf{Full-Profile} & \textbf{Rand. Occlus.} & \textbf{Rand. Act.} & \textbf{Average} \\
\hline\hline
VGG-Face  \cite{pvz15}        &2015 & 90.57\%           & 38.68\%      & 24.09\%       & 58.18\%          & 40.25\%       & 50.35\% \\
LFLBP  \cite{icip}           &2017 &  43.40\%           & 18.87\%      & 6.92\%       & 21.38\%          & 13.52\%       & 20.81\% \\
SE-ResNet-50 \cite{csxpz18} &      2018     & 89.30\%  & 79.87\% & 44.65\% & 72.64\% & 64.77\% & 70.25\%            \\
ResNet-50 \cite{csxpz18} &      2018     & 92.13\%  & 87.42\% & 53.45\% & 81.44\% &  74.84 \% &77.86\%          \\
LFHG \cite{ear}            &2018 &  38.99\%           & 6.92\%       & 3.14\%       & 23.58\%          & 18.24\%       & 18.17\% \\
VGG-D3  \cite{MLSP}          &2018 &  90.25\%           & 38.68\%      & 24.09\%       & 58.49\%          & 41.28\%       & 51.66\% \\
{ResNet50+ST-LSTM} \cite{LSTMV1}& 2018 & 92.14\%  &  86.79\% &  61.32\%  &  87.11\%  &    77.36\%    &  80.95\%   \\
VGG + Conv-LSTM  \cite{CSVT} &2019 & 98.11\%           & 78.62\%      & 22.01\%      & 81.13\%          & 71.38\%       & 70.25\% \\
{ArcFace} \cite{arc}& 2019 &  94.34\% &  84.91\%  &  51.57\%  &  80.50\%  &  70.44\%    &  76.35\%      \\
VGG + GLF-LSTM  \cite{joint}  &2020 & \textbf{99.37}\%           & 80.82\%      & 15.41\%      & 83.96\%          & 69.18\%       & 69.75\% \\
VGG + SLF-LSTM  \cite{joint}  &2020 & 98.74\%           & 77.36\%      & 22.64\%      & 80.82\%          & 67.92\%       & 69.50\% \\
{ResNet50+DS-LSTM} \cite{LSTMV2}& 2020 &  90.25\% &  84.28\%  &  60.69\%  &  85.85\%  &  76.10\%    &  79.43\%    \\

\hline
Prop. ResNet50 + Capsule   & ---      & 96.22\%           & \textbf{92.13}\%      & \textbf{67.29}\%      & \textbf{89.93}\%          & \textbf{85.22}\%       & \textbf{86.16}\%\\
\hline
\end{tabular}
\label{tab:pose}
\end{table*}

\begin{table*}[]
\centering
\setlength\tabcolsep{4.5pt}
\footnotesize
\centering
\caption{Cross-dataset protocol assessment: Face recognition rank-1 results using LFFC for training and LFFW for testing.}
\begin{tabular}{l|l|llllll|l}
\hline
\textbf{Solution}     & \textbf{Year}                & \textbf{Neutral} & \textbf{Rand. Exp.} & \textbf{Half-Profile} & \textbf{Full-Profile} & \textbf{Rand. Occl.} & \textbf{Rand. Act.} & \textbf{Average} \\
\hline\hline
VGG-Face \cite{pvz15}   & 2015     & 91.51\%         & 85.85\%           & 51.89\%      & 11.01\%      & 57.23\%          & 46.86\%       & 57.38\% \\
LFLBP \cite{icip}     & 2017     & 11.64\%         & 13.21\%           & 10.37\%      & 5.66\%       & 8.49\%           & 6.60\%        & 9.33\%  \\
SE-ResNet-50 \cite{csxpz18} &      2018     & 78.30\%  & 77.35\% & 69.18\% & 51.88\% & 66.03\% & 58.80\% & 66.92\%          \\
ResNet-50 \cite{csxpz18} &      2018     & 87.73\%  &  86.16\%  & 77.67\%  & 52.51\% & 73.27\% & 67.29\% & 74.10\%          \\
LFHG \cite{ear}       & 2018     & 30.19\%         & 27.99\%           & 11.01\%      & 7.55\%       & 12.58\%          & 8.18\%        & 16.24\% \\
VGG-D3 \cite{MLSP}     & 2018     & 90.25\%         & 85.22\%           & 51.26\%      & 10.69\%      & 59.75\%          & 44.97\%       & 57.02\% \\
{ResNet50+ST-LSTM} \cite{LSTMV1}& 2018 &  \textbf{93.39}\% & 90.57\% &  81.76\% &  \textbf{61.64}\% &  \textbf{82.07}\% &  73.90\% &  80.56\% \\ 
VGG + Conv-LSTM \cite{CSVT} & 2019 & 90.57\%         & 86.16\%           & 71.07\%      & 34.48\%      & 72.01\%          & 60.69\%       & 69.50\% \\
{ArcFace} \cite{arc}& 2019 &  91.51\% &  88.05\%  &  72.01\%  &  47.80\%  &  74.84\%    &  64.15\%  &    73.06\%\\
VGG + GLF-LSTM  \cite{joint}  & 2020 &  89.94\%         & 83.33\%           & 72.96\%      & 41.82\%      & 72.33\%          & 57.55\%       & 69.65\% \\
VGG + SLF-LSTM  \cite{joint}  & 2020 & 91.51\%         & 83.65\%           & 71.07\%      & 40.88\%      & 71.07\%          & 58.81\%       & 69.50\% \\
{ResNet50+DS-LSTM} \cite{LSTMV2}& 2020 &  92.76\% & 89.62\% &  79.87\% & 60.37\% &  80.81\% &  73.58\% &  79.51\%   \\

\hline
Prop. ResNet50 + Capsule    & ---     & 92.76\%         & \textbf{90.88}\%           & \textbf{82.70}\%      & 61.32\%      & \textbf{82.07}\%          & \textbf{74.84}\%       & \textbf{80.76}\%\\
\hline
\end{tabular}
\label{tab:DB1}
\end{table*}

\begin{table*}[]
\centering
\setlength\tabcolsep{5pt}
\footnotesize
\centering
\caption{Cross-dataset protocol assessment: Face recognition rank-1 results using LFFW for training and LFFC for testing.}
\begin{tabular}{l|l|llllll|l}
\hline
\textbf{Solution}                     & \textbf{Year} & \textbf{Neutral} & \textbf{Exp.} & \textbf{Action}  & \textbf{Pose}    & \textbf{Illum.} & \textbf{Occlus.} & \textbf{Average} \\
\hline\hline
VGG-Face \cite{pvz15}       & 2015 & 84.91\%         & 85.53\% & 84.91\% & 68.87\% & 82.08\%      & 67.92\%   & 74.81\% \\
LFLBP \cite{icip}           & 2017 & 17.92\%         & 13.21\% & 16.98\% & 15.09\% & 13.21\%      & 17.92\%   & 15.72\% \\
SE-ResNet-50 \cite{csxpz18} &      2018     &  83.01\% & 79.87\% & 77.35\% & 81.76\% & 85.84\% & 69.81\%& 77.92\%          \\
ResNet-50 \cite{csxpz18} &      2018     & 96.22\%  & 94.96\% & 92.45\% & 86.62\% & 96.22\% & 82.38\% & 89.52\%            \\
LFHG \cite{ear}           & 2018 & 27.35\%         & 34.90\% & 25.47\% & 18.86\% & 33.01\%      & 16.03\%   & 25.94\% \\
VGG-D3 \cite{MLSP}          & 2018 & 84.91\%         & 86.16\% & 83.96\% & 69.18\% & 83.02\%      & 68.24\%   & 75.09\% \\
{ResNet50+ST-LSTM} \cite{LSTMV1}& 2018 &  98.11\% & 96.23\% &  92.45\% &  94.65\% &  97.17\% &  82.28\% &  91.98\%  \\
VGG + Conv-LSTM  \cite{CSVT} & 2019 & 92.45\%         & 89.94\% & 90.57\% & 76.10\% & 88.68\%      & 73.27\%   & 80.85\% \\
{ArcFace} \cite{arc}& 2019 &  98.11\% &  \textbf{98.11}\%  &  \textbf{97.16}\%  &  78.93\%  &  97.16\%    &  \textbf{90.88}\%  &  90.00\%   \\
VGG + GLF-LSTM  \cite{joint}  & 2020 & 90.57\%         & 90.57\% & 88.68\% & 78.30\% & 89.62\%      & 71.07\%   & 80.75\% \\
VGG + SLF-LSTM  \cite{joint}  & 2020 & 88.68\%         & 89.94\% & 89.62\% & 79.25\% & 87.74\%      & 71.70\%   & 80.94\% \\
{ResNet50+DS-LSTM} \cite{LSTMV2}& 2020 &  96.27\% & 96.86\% & 93.40\% &  93.71\% &  97.17\% &  82.70\% &  91.32\%  \\

\hline
Prop. ResNet50 + Capsule         & --- & \textbf{100}\%           & \textbf{98.11}\% & 96.22\% & \textbf{97.48}\% & \textbf{100}\%        & 90.56\%   & \textbf{95.75}\%\\
\hline
\end{tabular}
\label{tab:DB2}
\end{table*}

\begin{table*}[]
\centering
\setlength\tabcolsep{4.5pt}
\footnotesize
\centering
\caption{{Subject-independent protocol assessment: Face recognition rank-1 results.}}
\begin{tabular}{l|l|lllll|l}
\hline
\textbf{Solution}  & \textbf{Year} &      \textbf{Rand. Exp.} & \textbf{Half-Profile} & \textbf{Full-Profile} & \textbf{Rand. Occlus.} & \textbf{Rand. Act.} & \textbf{Average} \\
\hline\hline

VGG-Face  \cite{pvz15}        &2015 & 88.00\%           & 32.00\%      & 24.67\%       & 57.33\%          & 31.33\%       & 50.35\%\\

LFLBP  \cite{icip}           &2017 &  40.67\%           & 20.67\%      & 06.00\%       & 22.67\%          & 12.00\%       & 20.81\% \\
SE-ResNet-50 \cite{csxpz18} &      2018     & 91.33\%           & 81.33\%      & 49.33\%       & 81.67\%          & 72.67\%       & 75.26\% \\
ResNet-50 \cite{csxpz18} &      2018      & 90.67\%           & 82.67\%      & 50.67\%       & 81.33\%          & 74.00\%       & 75.87\% \\
LFHG \cite{ear}            &2018 & 40.67\%           & 08.00\%      & 2.67\%       & 22.00\%          & 17.33\%       & 18.18\% \\
VGG-D3  \cite{MLSP}          &2018 & 89.33\%           & 40.00\%      & 24.67\%       & 63.33\%          & 40.00\%       & 51.47\% \\
ResNet50+ST-LSTM \cite{LSTMV1}& 2018 &  92.00\% &  \textbf{84.67}\%  &  \textbf{55.67}\%  &  84.33\%  &  74.00\%    &  78.13\%    \\
VGG + Conv-LSTM  \cite{CSVT} &2019 &  89.33\%           & 71.33\%      & 43.33\%       & 72.67\%          & 60.67\%       & 67.46\% \\
ArcFace \cite{arc}& 2019 &  \textbf{95.33}\% &  77.33\%  &  47.33\%  &  \textbf{88.00}\%  &  72.00\%    &  76.00\%   \\
VGG + GLF-LSTM  \cite{joint}  &2020 & 89.33\%           & 71.33\%      & 46.66\%       & 78.00\%          & 62.67\%       & 69.59\% \\
VGG + SLF-LSTM  \cite{joint}  &2020 & 88.67\%           & 73.33\%      & 42.00\%       & 76.00\%          & 60.00\%       & 68.00\% \\
ResNet50+DS-LSTM \cite{LSTMV2}& 2020 &  91.67\% &  \textbf{84.67}\%  &  53.33\%  &  85.33\%  &  74.67\%    &  77.93\%    \\

\hline
Prop. ResNet50 + Capsule   & ---      &  93.33\%           &  84.00\%      &   54.67\%      &  86.67\%          &  \textbf{76.00}\%       &  \textbf{78.93}\%\\
\hline
\end{tabular}
\label{tab:ind}
\end{table*}

\subsection{Comparative Face Expression Recognition Performance}
This section reports the expression recognition accuracy obtained for the two proposed evaluation protocols. The performance results are reported in Table \ref{tab:exp} for the proposed and 6 benchmarking solutions, with the best results highlighted in bold. 


The performance results for both evaluation protocols clearly show that the proposed CapsField solution achieves considerably better performance than the non-LF-based deep recognition solutions, presented in the first four rows of Table \ref{tab:exp}. This improvement is due to the fact that the proposed solution exploits the available spatio-angular information in the LFs for the face expression recognition task. The results also show the superiority of the proposed CapsField solution over the LF-based ones, notably VGG+LSTM \cite{ACII} and VGG+BiLSTM+Attention \cite{ICASSP} solutions. As the comparative LF-based and the proposed solutions use the same spatial CNN model, i.e., VGG-16 pre-trained with VGG-Face 2, the performance results reveal that the capsule network better models the richer angular features than the LSTM and BiLSTM recurrent networks used in the benchmarking solutions for the LF expression recognition task.

\subsection{{Benefits of LF Images for Face Recognition}}
{An additional experiment has been performed to evaluate the added value of LF images in comparison with single-view images for face recognition. This considers the comparison between four LF-based face recognition methods including LFLBP \cite{icip}, HOG+HDG \cite{ear}, VGG-16 + LSTM \cite{CSVT}, and the proposed ResNet-50 + Capsule, and their corresponding single-view versions, including LBP \cite{LBP}, HOG \cite{HOG}, VGG-16 \cite{pvz15}, and ResNet-50 \cite{csxpz18}, applied to the central SA image. The results presented in Table \ref{tab: LFres} clearly demonstrate the superiority of the LF-based methods over their corresponding single-view baselines for the four subject-dependent test protocols, thus highlighting the added value of LF images for face recognition.}

\begin{table}[!t]
\centering
\footnotesize
\centering
\caption{Expression recognition accuracy using two different evaluation protocols.}
\setlength
\tabcolsep{3pt}
\begin{tabular}{ l| l| l| l| l}
\hline
\textbf{Solution} & \textbf{Year} & \textbf{4-Fold} & \textbf{Cross-} & \textbf{Average} \\ 
&  & \textbf{CV} & \textbf{DS}&   \\ 
\hline\hline

    VGG-16 (VGG-Face2 model) \cite{csxpz18} & 2018 &  59.37\% & 51.34\%& 55.35\%\\
    VGG-16+ST-LSTM \cite{LSTMV1}& 2018 & 63.33\% & 52.80\% & 58.07\%\\
    VGG-16+LSTM \cite{ACII}& 2019 & 59.37\% & 51.10\%& 55.23\%\\
    VGG-16+BiLSTM+Attention \cite{ICASSP}& 2020 & 63.33\% & 52.90\% & 58.11\%\\
    VGG-16+DS-LSTM \cite{LSTMV2}& 2020 & 64.17\% & 56.29\% & 60.23\%\\
    Proposed VGG-16 + Capsule & --- & \textbf{65.74\%}& \textbf{57.45\%}&  \textbf{61.59\%} \\

    \hline
\end{tabular}
\label{tab:exp}
\end{table}

\begin{table}[]
\centering
\caption{{Face recognition performance for selected conventional 2D baseline solutions against their LF-based variants.}}
\setlength
\tabcolsep{2.5pt}
\begin{tabular}{l|l|l|l|l}
\hline
Solution          & \textbf{Cross-Env.} & \textbf{Cross-Dis.} & \textbf{Cross-Pose}  & \textbf{Cross-DS}\\

\hline
\hline
LBP \cite{LBP}              & 06.24\%  & 10.95\%  & 19.11\%  &  10.27\%  \\
LFLBP \cite{icip}           & 07.81\%  & 16.26\%  & 20.81\%  &  12.52\%  \\
\hline
HOG \cite{HOG}              & 20.65\%  & 21.96\%  & 17.73\%  &  19.06\%  \\
HOG+HDG \cite{ear}          & 23.85\%  & 24.92\%  & 18.17\%  &  21.09\%  \\
\hline
VGG-16 \cite{pvz15}         & 72.27\%  & 73.58\%  & 46.35\%  &  66.09\%  \\
VGG-16+LSTM \cite{CSVT}     & 85.80\%  & 85.45\%  & 70.25\%  &  75.17\%  \\
\hline
ResNet-50 \cite{csxpz18}    & 84.17\%  & 81.84\%  & 77.86\%  &  81.81\%  \\
Prop. ResNet50+Caps         & 87.83\%  & 89.10\%  & 86.16\%  &  88.25\%  \\
\hline
\end{tabular}
\label{tab: LFres}
\end{table}

\subsection{Angular Feature Space Exploration}
\begin{figure}[!t]
\centering
\includegraphics[width=0.93\columnwidth]{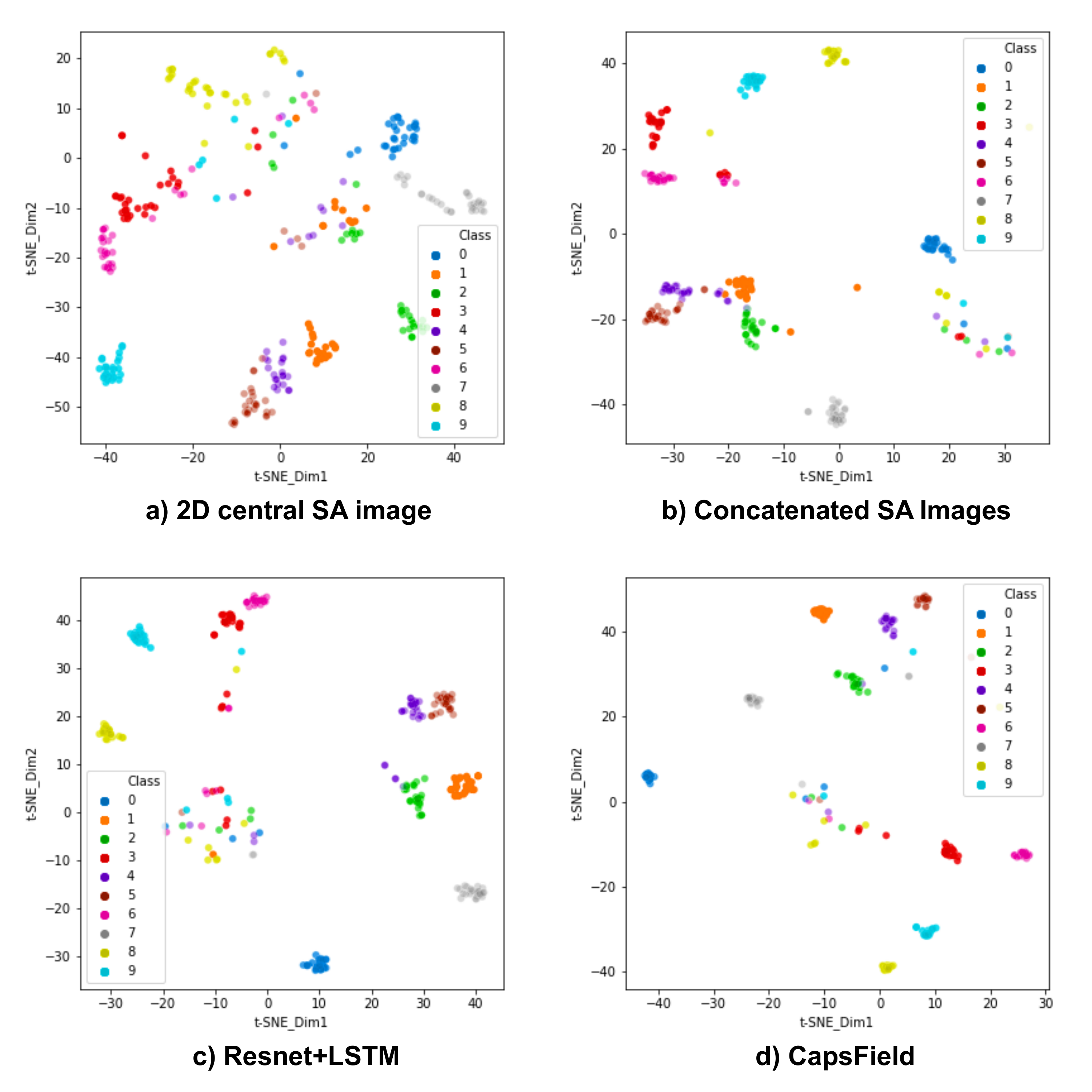}
\caption{t-SNE visualization of the feature spaces using four different angular exploitation strategies.}
\label{fig:tSne}
\end{figure}

To study the impact of different strategies for dealing with the available LF angular information, t-SNE \cite{tSNE} is used again to plot the feature spaces produced by the different strategies, thus showing the different discrimination ability of the various LF angular features in a two dimensional space. Figure \ref{fig:tSne} offers a visualization of the features that are used as inputs to the classifier, when four different strategies are used to deal with the angular information, notably: i) only the 2D central SA image is used, thus ignoring the available angular information (Figure \ref{fig:tSne}-a) ; ii) the concatenated features extracted from all the views are used (without capsule network) to feed the classifier (Figure\ref{fig:tSne}-b); iii) LSTM RNN \cite{CSVT} is used to exploit the angular information (Figure\ref{fig:tSne}-c); and iv) capsule network, as proposed in this paper, is used to exploit the angular information (Figure \ref{fig:tSne}-d). To make the visualisation legible, the t-SNE analysis is only performed for the first 10 LFFW subjects using the cross-pose and expression face recognition protocol. The t-SNE plots show that the proposed CapsField solution forms denser and more effective clusters that can facilitate distinguishing between the various classes; this validates the superiority of the capsule network in learning angular information.

\subsection{Ablation Study}
Ablation experiments have been performed based on the proposed evaluation protocols for the adopted datasets by systematically removing individual components of CapsField one by one and evaluating the performance of the 'reduced' models. It must be noted that the CNN sub-network must always be kept in order to extract the spatial features.
Table \ref{tab:abl} presents the ablation experiment results for both the face and expression recognition tasks. For comparison, the performance of the complete CapsField model is also presented in the first row of the table, where all components are used. The following subsections describe each ablation test.

\subsubsection{Impact of Horizontal and Vertical LF Views} To investigate the effectiveness of the fusion strategy, the individual contributions of the horizontal or vertical SA image sequences have been investigated. The results show that by combining the classification scores of both LF sequences, average performance gains of 0.80\% and 1.04\% are obtained, respectively, when compared to a CapsField version considering only the horizontal or vertical SA image sequences.

 
\subsubsection{Impact of Capsule Sub-network} Finally, the capsule component is removed, and the CNN spatial features obtained from different LF SA images are directly concatenated for classification. It is worth noting that the SE-ResNet-50 \cite{csxpz18} and VGG-16 \cite{csxpz18} results presented in Tables \ref{tab:env} through \ref{tab:DB2} are applied to a single 2D central SA image, while the features extracted from all the views are concatenated here. Removing the capsule sub-network
decreases the recognition performance of the proposed CapsField solution as it limits the ability to learn the inter-view/angular relations between the spatial features. Moreover, in the absence of the capsule component, it is not possible to assign higher weights to the more relevant features. These observations reveal the added value of the capsule sub-network in learning a model that fully exploits the angular features available in LF images. 

\begin{table}[!t]
\centering
\footnotesize
\centering
\caption{Ablation study for the proposed solution.}
\setlength
\tabcolsep{2pt}
\begin{tabular}{ l| l| l| l| l| l| l| l| l}
\hline
    \multicolumn{3}{ c| }{\textbf{Component}}& \multicolumn{4}{ c |}{\textbf{Face Rec. Protocol}}& \multicolumn{2}{ c }{\textbf{Exp. Rec. Protocol}}   \\ 
\hline

    \textbf{Hor.}& \textbf{Ver.}& \textbf{Caps}& \textbf{Cross-}& \textbf{Cross-}& \textbf{Cross-}& \textbf{Cross-}& \textbf{4-Fold}& \textbf{Cross-}\\
     &  &  & \textbf{Env.}& \textbf{Dist.}& \textbf{Pose}& \textbf{DS}& \textbf{CV}& \textbf{DS}

    \\ \hline\hline
    \cmark & \cmark &  \cmark &\textbf{87.83\%}  & \textbf{89.10\%} & \textbf{86.16\%} &  \textbf{88.25\%} &  \textbf{65.74\%} & \textbf{57.45\%}\\
    \xmark & \cmark &  \cmark &  86.47\% & 88.30\%  & 85.35\%  & 87.46\%  & 64.53\% & 56.13\%\\
    \cmark & \xmark &  \cmark & 87.15\%  &  88.36\% & 84.53\%  & 87.34\%  &  65.14\% & 57.27\%\\
    \cmark & \cmark &  \xmark & 85.65\%  &  85.50\% &  83.83\% & 82.13\%  & 61.85\% & 53.75\% \\
    \hline
\end{tabular}
\label{tab:abl}
\end{table}

\subsection{{Complexity Analysis}}

{The computational time and embedding size for the proposed and benchmarking recognition solutions are studied in this section. This analysis has been done by measuring the execution times on a 64-bit Intel PC with a 3.20 GHz Core i7 processor, 48 GB RAM, and a GeForce GTX 1080 Ti GPU, running TensorFlow with Keras backend. Table \ref{tab:time} shows the training and testing times (in seconds) per each 2D/LF image as well as the final embedding size in terms of the number of embedding elements. It can be observed from Table \ref{tab:time} that the proposed CapsField solution offers the most compact embedding when compared to the benchmarking solutions, thus simplifying the retrieval and transmission of the embeddings as well as reducing the computational time for the testing phase. The required training time for the CapsField solution is higher than other benchmarking solutions, which is the trade-off for extracting more discriminative features. It is also worth noting that the very high training and testing times for the VGG-D3 \cite{MLSP} solution derives from the necessary disparity and depth maps extraction processes that are computationally very expensive.}

\begin{table}[]
\centering
\setlength\tabcolsep{5pt}
\footnotesize
\centering
\caption{{Average training and testing times per image (in seconds), and embedding size (number of elements) for the proposed and benchmarking face recognition solutions.}}
\begin{tabular}{l|l|lll}
\hline
\textbf{Solution}     & \textbf{Type}                & {\vtop{\hbox{\strut \textbf{Training}}\hbox{\strut \textbf{Time}}}}  & {\vtop{\hbox{\strut \textbf{Testing}}\hbox{\strut \textbf{Time}}}} & {\vtop{\hbox{\strut \textbf{Embedding}}\hbox{\strut \textbf{Size}}}}\\
\hline\hline
VGG-Face \cite{pvz15}         & 2D     & 0.045 & 0.035  & 4,096  \\
SE-ResNet-50 \cite{csxpz18}    & 2D    & 0.032 & 0.027 & 2,048  \\
ResNet-50 \cite{csxpz18}       & 2D    & 0.031 & 0.027 & 2,048  \\
ArcFace \cite{arc}             & 2D    & 0.072 & 0.023 & 512   \\
LFLBP \cite{icip}             & LF     & 0.466 & 0.276 & 65,536 \\
LFHG \cite{ear}               & LF     & 0.667 & 0.577 & 16,200 \\
VGG-D3 \cite{MLSP}            & LF     & 397.657 & 397.643 & 16,384 \\
VGG + Conv-LSTM \cite{CSVT}   & LF     & 0.808 & 0.513 & 7,680  \\
VGG + ST-LSTM  \cite{LSTMV1}  & LF     & 1.183 & 0.497 & 3,840  \\
VGG + SLF-LSTM  \cite{joint}  & LF     & 1.003 & 0.497 & 3,840  \\
VGG + GLF-LSTM  \cite{joint}  & LF     & 1.002 & 0.497 & 3,840  \\
VGG + DS-LSTM  \cite{LSTMV2}  & LF     & 0.979 & 0.497 & 3,840  \\
\hline
Prop. ResNet50 + Capsule      & LF     & 1.971 & 0.488 & 320 \\
\hline
\end{tabular}
\label{tab:time}
\end{table}

\section{Conclusion}
This paper proposes a new solution for both face and expression recognition tasks, called CapsField, based on the combination of convolutional neural and capsule networks. In order to analyze the performance of CapsField in the wild, this paper proposes the first unconstrained LF face dataset, named LF Faces in the Wild (LFFW), affected by variations in resolution, background, expression, pose, illumination, and occlusions. This paper also proposes a LF Face Constrained (LFFC) dataset, complementary to the LFFW dataset that has been captured in a controlled environment, from the same subjects available in LFFW. A subset of the in the wild dataset, including the facial images with different expressions, has been annotated for face expression recognition. An extensive benchmarking study has been performed for challenging evaluation protocols on the developed datasets. The performance results have shown the superiority of the CapsField solution, compared to the state-of-the-art in LF-based face and expression recognition, due to its ability in learning inter-view  and  intra-view  relations  available  in  an  LF image.

\section{Acknowledgment}
This work is partly funded by FCT/MCTES under the project UIDB/50008/2020.

\bibliographystyle{ieeetran}
\bibliography{references}

\begin{IEEEbiography}[
{
\includegraphics[width=1in,height=1.25in,clip,keepaspectratio]{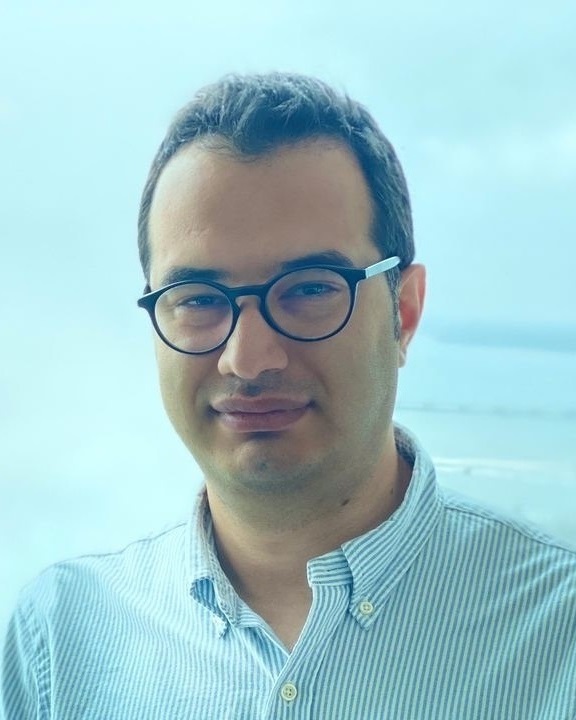}
}
]
{Alireza Sepas-Moghaddam}
received the B.Sc. and M.Sc. (first class honors) degrees in Computer Engineering in 2007 and 2010, respectively. From 2011 to 2015, he was with the Shamsipour University, Tehran, Iran, as a lecturer. In 2015, he joined Instituto Superior Técnico, University of Lisbon, Portugal, where he completed his Ph.D. degree with distinction and honour in Electrical and Computer Engineering in 2019. He is currently a Postdoctoral Fellow at the Department of Electrical and Computer Engineering, Queen’s University, Canada, where he works on different research projects funded by the Natural Sciences and Engineering Research Council of Canada and Mitacs, as well as private sector. His main research interests are machine learning and deep learning for biometrics, forensics, affective computing, and computer vision. He has contributed more than 40 papers in notable conferences and journals in his area and has been a reviewer for multiple top-tier conferences and journals. 
\end{IEEEbiography}

\begin{IEEEbiography}[
{
\includegraphics[width=1in,height=1.20in,clip,keepaspectratio]{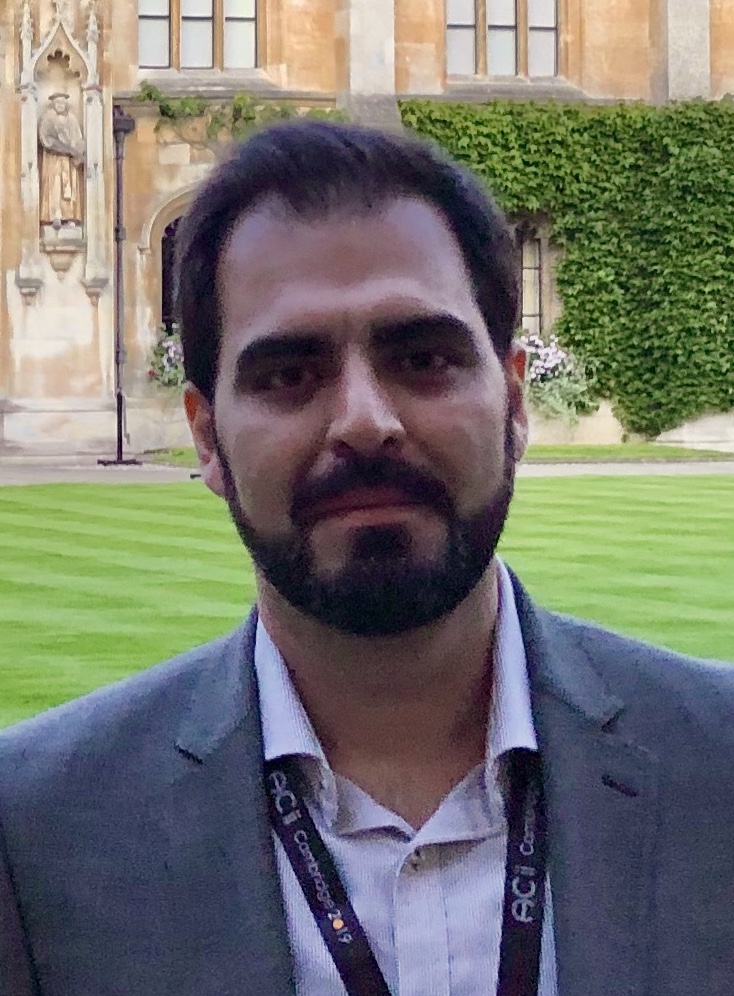}
}
]
{Ali Etemad} received the M.A.Sc. and Ph.D. degrees in Electrical and Computer Engineering from Carleton University, Ottawa, Canada, in 2009 and 2014, respectively. He is currently an Assistant Professor at the Department of Electrical and Computer Engineering, Queen’s University, Canada, where he leads the Ambient Intelligence and Interactive Machines (Aiim) lab. He is also a faculty member at Ingenuity Labs Research Institute. His main area of research is machine learning and deep learning focused on human-centered applications with wearables, smart devices, and smart environments. Prior to joining Queen’s, he held several industrial positions as lead scientist and director. He has co-authored over 90 articles and patents published/granted or under review, and has delivered over 15 invited talks regarding his work. He has been a member of program committees for several conferences in the field. He has been the recipient of a number of awards and grants. Dr. Etemad’s lab and research program have been funded by the Natural Sciences and Engineering Research Council of Canada (NSERC), Ontario Centers of Excellence (OCE), Canadian Foundation for Innovation (CFI), Mitacs, and other organizations, as well as the private sector.

\end{IEEEbiography}

\begin{IEEEbiography}[
{
\includegraphics[width=1in,height=1.25in,clip,keepaspectratio]{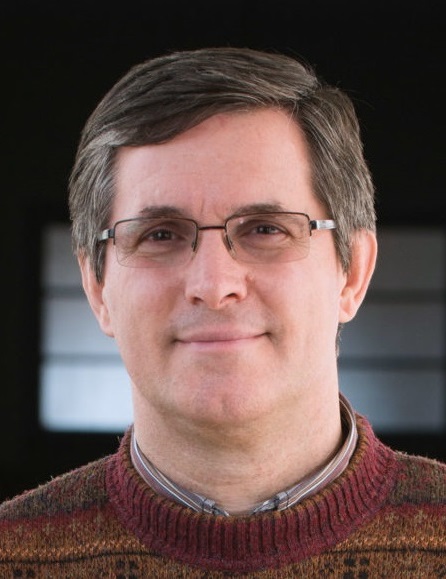}
}
]
{Fernando Pereira} (S’88–M’90–SM’99–F’08) is currently a Professor at the Department of Electrical and Computers Engineering, Instituto Superior Técnico, Universidade de Lisboa, and a Senior Researcher at Instituto de Telecomunicações, Lisbon, Portugal. He is Area Editor of the Signal Processing: Image Communication Journal and Associate Editor of the EURASIP Journal on Image and Video Processing, and is or has been member of the Editorial Board of the Signal Processing Magazine, Associate Editor of IEEE Transactions of Circuits and Systems for Video Technology, IEEE Transactions on Image Processing, IEEE Transactions on Multimedia, and IEEE Signal Processing Magazine. In 2013-2015, he was the Editor-in-Chief of the IEEE Journal of Selected Topics in Signal Processing. He was an IEEE Distinguished Lecturer in 2005 and elected as an IEEE Fellow in 2008 for “contributions to object-based digital video representation technologies and standards”. He has been elected to serve on the Signal Processing Society Board of Governors in the capacity of Member-at-Large for a 2012 and a 2014-2016 term. Since January 2018, he is the SPS Vice-President for Conferences. Since 2013, he is also a EURASIP Fellow for “contributions to digital video representation technologies and standards”. Since 2015, he is also a IET Fellow. He is/has been a member of the Scientific and Program Committees of many international conferences and workshops. He has been participating in the MPEG standardization activities, notably as the head of the Portuguese delegation, MPEG Requirements Subgroup Chair, and chair of many Ad Hoc Groups related to the MPEG-4 and MPEG-7 standards. Since February 2016, he is the JPEG Requirements Subgroup Chair. He has contributed more than 300 papers in international journals, conferences and workshops, and made several tens of invited talks at conferences and workshops. His areas of interest are visual data analysis, coding, description, adaptation, quality assessment, and advanced multimedia services. 

\end{IEEEbiography}

\begin{IEEEbiography}[
{
\includegraphics[width=1in,height=1.25in,clip,keepaspectratio]{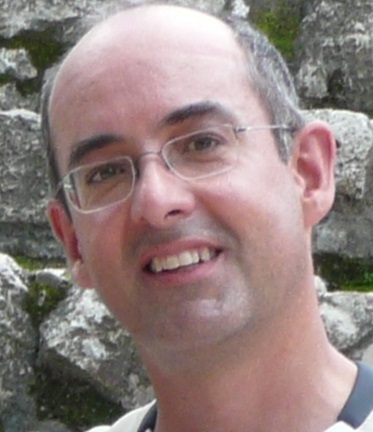}
}
]
{Paulo Lobato Correia} (M’05–SM’08) obtained the Engineering, MSc and PhD degrees in Electrical and Computer Engineering from Instituto Superior Técnico (IST), Universidade de Lisboa (UL), Portugal, in 1989, 1993 and 2002, respectively. He is Associate Professor at IST-UL and senior researcher and group leader at Instituto de Telecomunicações. He coordinated the participation in several national and international research projects dealing with image and video analysis and processing. He is Editor in Chief for IET Biometrics (2020-2022), Subject Editor (for Multimedia papers) of the Elsevier Signal Processing Journal (2018-2020). He was associate editor of the IEEE Transactions on Circuits and Systems for Video Technology (2006-2014), of the Elsevier Signal Processing Journal (2005-2017) and IET Biometrics (2013-2019). He has been guest editor of several special issues of scientific journals and cooperated in many conference organizing committees, as general chair, program chair, finance chair or special sessions chair. He is a founding member of the Advisory Board of EURASIP, and he is the elected chairman of EURASIP's Technical Area Committee on Biometrics, Data Forensics and Security. He has co-authored more than 135 conference and journal papers. Research interests include image and video analysis and processing, with emphasis on biometric recognition. 

\end{IEEEbiography}

\end{document}